\documentclass[lettersize,journal]{IEEEtran}
\usepackage{amsmath,amsfonts}
\usepackage{array}
\usepackage[caption=false,font=normalsize,labelfont=sf,textfont=sf]{subfig}
\usepackage{textcomp}
\usepackage{stfloats}
\usepackage{url}
\usepackage{verbatim}
\usepackage{graphicx}
\usepackage{cite}

\usepackage{adjustbox}
\usepackage{array}
\usepackage{booktabs}
\usepackage{multirow}
\usepackage{diagbox}

\newcolumntype{R}[2]{%
    >{\adjustbox{angle=#1,lap=\width-(#2)}\bgroup}%
    l%
    <{\egroup}%
}
\usepackage{amsmath, amssymb}
\DeclareMathOperator*{\argmin}{arg\,min}
\usepackage{graphics} % for pdf, bitmapped graphics files
\usepackage{epsfig} % for postscript graphics files
\usepackage{times} % assumes new font selection scheme installed
\usepackage{amsmath} % assumes amsmath package installed
\usepackage{amssymb}  % assumes amsmath package installed
\usepackage[table]{xcolor}
\usepackage[ruled,linesnumbered, noend]{algorithm2e}
\SetKwRepeat{Do}{do}{while}
\usepackage{adjustbox}

\usepackage{soul}
\usepackage{cite}
\usepackage{comment}
\usepackage{multirow}
\usepackage{wrapfig}
\usepackage{balance}
\usepackage{capt-of}% or \usepackage{caption}
\newcommand{\ve}[1]{\mathbf{#1}} % vector
\newcommand{\tve}[1]{\tilde{\mathbf{#1}}} % vector
\begin{document}

\title{Learning Haptic-based Object Pose Estimation\\ for In-hand Manipulation Control \\with Underactuated Robotic Hands}

\author{Osher~Azulay, Inbar Ben-David and Avishai~Sintov
\thanks{O. Azulay, I. Ben-David and A. Sintov are with the School of Mechanical Engineering, Tel-Aviv University, Israel. e-mail: {\small osherazulay@mail.tau.ac.il, \small sintov1@tauex.tau.ac.il}}
\thanks{This work was supported by the Pazy Foundation (grant No. 283-20).}
}

% \author{IEEE Publication Technology,~\IEEEmembership{Staff,~IEEE,}
%         % <-this % stops a space
% \thanks{This paper was produced by the IEEE Publication Technology Group. They are in Piscataway, NJ.}% <-this % stops a space
% \thanks{Manuscript received April 19, 2021; revised August 16, 2021.}}

% The paper headers
% \markboth{IEEE Transactions on Haptics,~Vol.~, No.~}%
% {Azulay \MakeLowercase{\textit{et al.}}: Learning Haptic-based Object Pose Estimation for In-hand Manipulation control with Underactuated Robotic Hands}

% \IEEEpubid{0000--0000/00\$00.00~\copyright~2021 IEEE}
% Remember, if you use this you must call \IEEEpubidadjcol in the second
% column for its text to clear the IEEEpubid mark.

\maketitle

\begin{abstract}
Unlike traditional robotic hands, underactuated compliant hands are challenging to model due to inherent uncertainties. Consequently, pose estimation of a grasped object is usually performed based on visual perception. However, visual perception of the hand and object can be limited in occluded or partly-occluded environments. 
%vision is often limited due to lighting or occlusion of the object by the fingers and environment. 
In this paper, we aim to explore the use of haptics, i.e., kinesthetic and tactile sensing, for pose estimation and in-hand manipulation with underactuated hands. Such haptic approach would mitigate occluded environments where line-of-sight is not always available. We put an emphasis on identifying the feature state representation of the system that does not include vision and can be obtained with simple and low-cost hardware. For tactile sensing, therefore, we propose a low-cost and flexible sensor that is mostly 3D printed along with the finger-tip and can provide implicit contact information. Taking a two-finger underactuated hand as a test-case, we analyze the contribution of kinesthetic and tactile features along with various regression models to the accuracy of the predictions. Furthermore, we propose a Model Predictive Control (MPC) approach which utilizes the pose estimation to manipulate objects to desired positions solely based on haptics. We have conducted a series of experiments that validate the ability to estimate poses of various objects with different geometry, stiffness and texture, and show manipulation to goals in the workspace with relatively high accuracy.
\end{abstract}

\begin{IEEEkeywords}
Pose estimation, in-hand manipulation, underactuated hands.
\end{IEEEkeywords}

\section{Introduction}
\label{sec:introduction}

% Robots require efficient interaction with a large variety of objects in different environments, from industrial applications to domestic tasks. The interaction involves grasping and dexterous manipulation of objects to complete various tasks. 

\IEEEPARstart{W}{hile} the ability to manipulate an object within the hand is a fundamental everyday task for humans, such problem remains challenging for robots. Traditional robotic hands, such as the Shadow and the Allegro hands \cite{Bae2012}, have achieved significant accuracy and performance. However, they have complex structure % with many degrees-of-freedom 
while being fragile, costly and difficult to control \cite{Dollar2010,Bai2014,Michalec2010}. \textit{Underactuated hands}, on the other hand, are mechanisms that can adapt to the shape of the object through the use of compliance \cite{Odhner2011}. Consequently, they have gained popularity in recent years due to their low-cost and ability to maintain a stable grasp with open-loop control. In addition, precision manipulation with underactuated hands have been shown possible \cite{Odhner2015,Calli2016,Calli2018}. However, certain limitations hinder practical usage. While pose estimation is generally available analytically in rigid hands where the kinematics are known \cite{Ozawa2005}, such model is rarely available for underactuated ones \cite{Sintov2019}. Acquiring a precise analytical model for an underactuated hand is not always easy or feasible due to inherent uncertainties and fabrication inaccuracies. Hence, estimating properties such as joint stiffness, size, weight, friction and inertia is a significant challenge \cite{Borras2013}. 
To cope with the lack of an analytical solution, data-based modeling was shown useful in providing accurate predictions in motion planning and control \cite{icra2020a,Belief2019,Morgan2021c}. Such modeling approach is able to intrinsically estimate model parameters that can be difficult or impossible to model otherwise.

%%%%%%%%%%%%%%%%%%%%%%%%%%%%%%%
 \begin{figure}[t!]
    \centering
    \includegraphics[width=\linewidth,keepaspectratio]{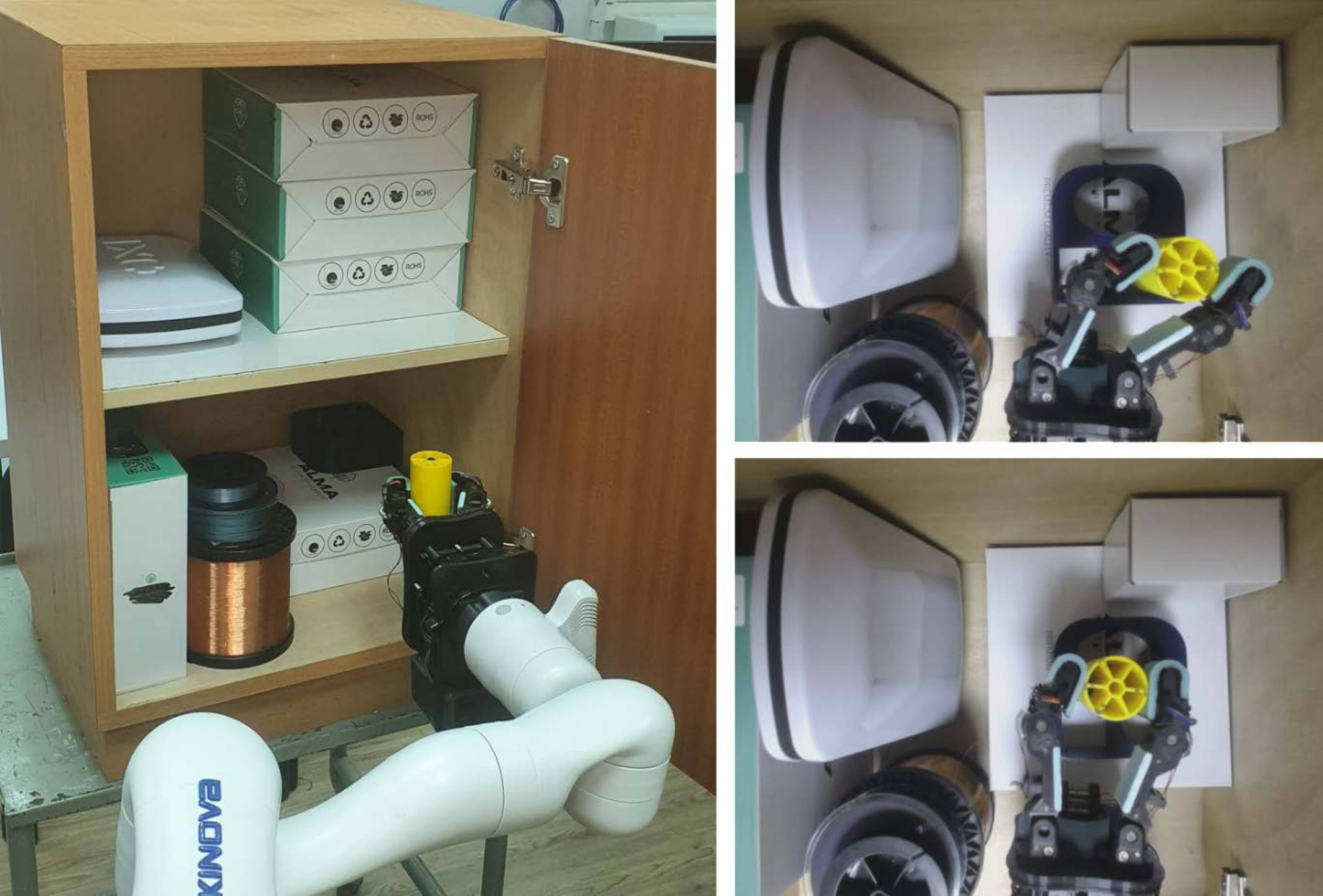}
    \caption{Manipulating an object within an occluded cabinet using a robotic arm and an underactuated hand into a target slot. Visual perception is not available within the cabinet and, therefore, the hand must use haptic perception. We assume that the position of the hole relative to the robot is known.}
    \label{fig:hand}
    \vspace{-0.4cm}
\end{figure}
%%%%%%%%%%%%%%%%%%%%%%%%%%%%%%%
As mentioned above, exact analytical solutions are rarely available for underactuated hands. Therefore, attempts to model these hands in the context of precision manipulation usually rely on external visual feedback. Calli and Dollar \cite{Calli2016} used a linear approximation of a two-finger hand to apply visual servoing to track desired paths. Recently, Morgan et al. \cite{Morgan2020} proposed an object agnostic manipulation using a vision-based Model Predictive Control (MPC). The work in \cite{Wen2020} used a depth camera to estimate the pose of an object grasped and partly-occluded by the two-fingers of an underactuated hand. Extension of the work proposed the use of the depth-based 6D pose estimation to control precise manipulation of a grasped object \cite{Morgan2021}. Recent work \cite{Fonseca2019} integrated allocentric visual perception along with four tactile modules, that combine pressure, magnetic, angular velocity and gravity sensors, on two underactuated fingers. These sensors were used to train a pose estimation model. More related, Sintov et al. \cite{Sintov2019} proposed a data-based transition model where the state of the hand involves kinesthetic features such as actuator torques and angles along with the position of the manipulated object acquired with visual feedback. While a visual approach may exhibit good results, relying on continuous visual feedback limits the performance of various tasks in which visual uncertainty (e.g., poor lighting or shadows) or occlusion may occur. In such cases, it may be impossible to solve the task altogether. This may include operating tools at the back of a cabinet or in confined spaces (Figure \ref{fig:hand}). 

Information regarding grasped objects are often acquired through haptic perception including both tactile sensing and internal sensing of joint actuators known as Kinesthetic haptics \cite{Carter2005}. Traditionally, tactile refers to information received from touch and contact sensing, while kinesthetic refers to information sensed through movement, force or position of joints and actuators. Tactile sensing %with or without visual perception 
is the leading method for haptic-based object recognition \cite{Rouhafzay2020,Pohtongkam2021,Liu2022}. State of the art in tactile sensing is focused on optical devices where an internal camera observes the deformation of a flexible surface during contact with an object \cite{Chorley2009,Donlon2018}. Recent work has explored the use of these optical tactile sensors with advanced deep networks to estimate the relative pose of an object in contact \cite{Lepora2020}. While these sensors can provide accurate performance, they can complicate the hardware.%, are not easy to use and require a large tactile dataset for training.
%%%%%%%%%%%%%%%%%%%%%%%
\begin{figure*}
\centering
% \begin{tabular}{cc}
% \includegraphics[width=0.5\linewidth,keepaspectratio]{figures/finger_finger_iso.png} & \hfill
% \includegraphics[width=0.48\linewidth,keepaspectratio]{figures/finger_profile.png}\\
% (a) & (b)
% \end{tabular}
% \includegraphics[width=\linewidth,keepaspectratio]{figures/tactile.PNG}
\includegraphics[width=\linewidth,keepaspectratio]{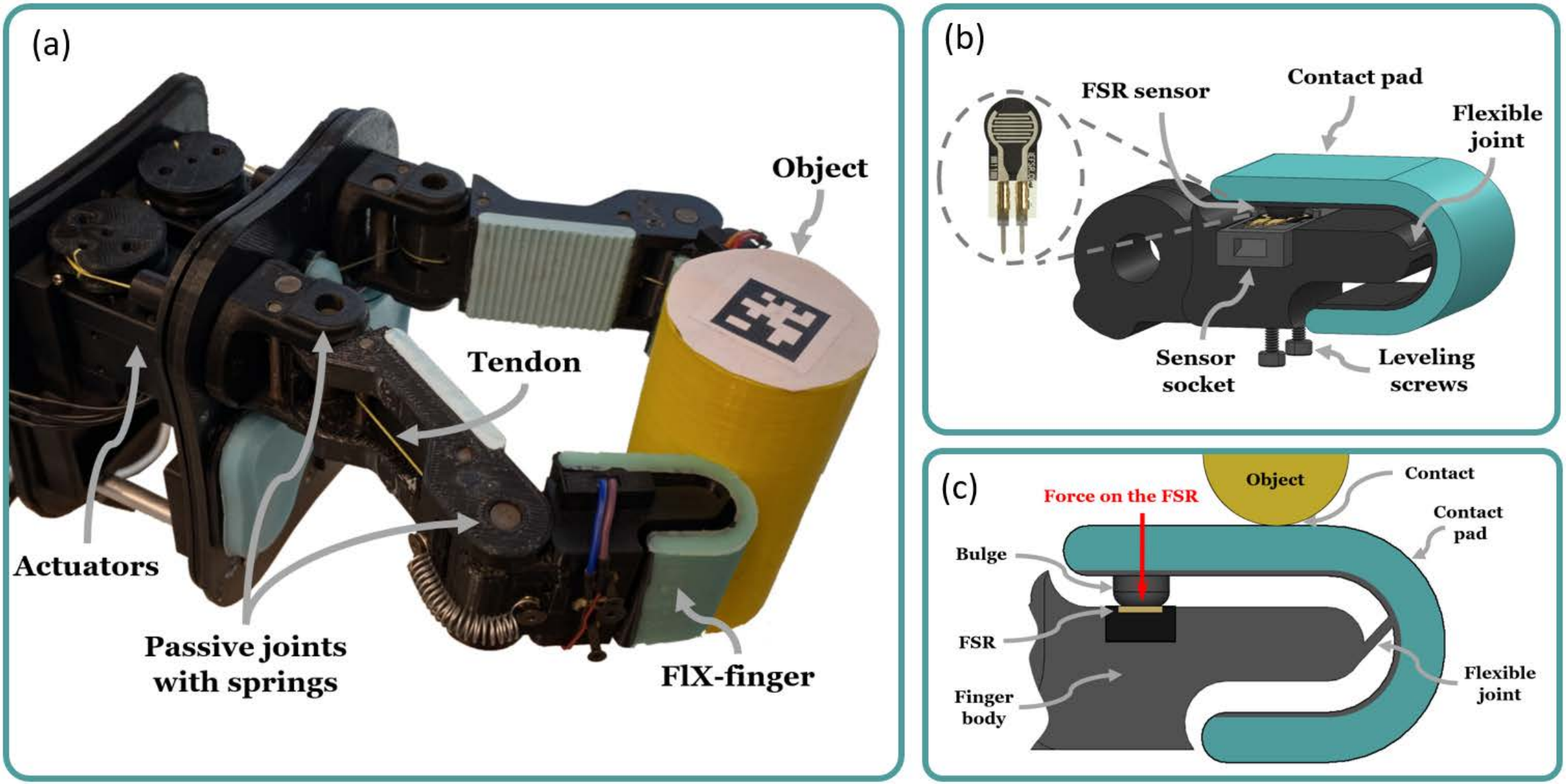}
\caption{(a) Structure of a tendon-based underactuated hand used in this work. The hand is comprised of two opposing fingers. Each finger has two passive joints with springs where a tendon wire runs along its length and is connected to an actuator. In addition, each finger includes a FlX-finger for contact sensing. A (b) detailed and (c) side view of the FlX-finger with the FSR sensor are also seen. When contact force is exerted on the contact pad by the object, the flexible joint and pad deform. Consequently, the bulge is pressed onto the FSR which, in turn, outputs a voltage that depends on the contact distance from the bulge across the pad along with the magnitude of the normal contact force. }
\label{fig:FlX-finger}
\vspace{-0.4cm}
\end{figure*}
%%%%%%%%%%%%%%%%%%%%%%%

The abilities of underactuated hands in various tasks and their advantages over traditional rigid hands have been widely demonstrated in prior work \cite{Dollar2010,Calli2016,Calli2017,Calli2018a,Sintov2019,icra2020a}. However and as discussed above, these abilities rely on extrinsic visual perception. In this paper, we investigate the ability to track and manipulate an object within a tendon-based two-finger underactuated hand (seen in Figure \ref{fig:FlX-finger}) solely based on haptics. As opposed to prior work where the hand's state included explicit object position acquired from vision \cite{Sintov2019}, visual perception is not relied upon in this work. Hence, we explore the sole use of kinesthetic information available to the hand along with novel low-cost tactile fingers termed \textit{FlX-fingers}. The FlX-finger is mostly 3D printed along with a simple off-the-shelf force Sensitive Resistor (FSR). While the use of FSR for tactile perception is not novel \cite{Kappassov2015}, we propose a flexible 3D printed mechanism that augments the sensor and provides more contact data. Therefore and unlike custom made force sensors (e.g., \cite{Muth2014}), we use a standard off-the shelf FSR sensor and a 3D printed finger structure to acquire diverse contact data. 

With the FlX-fingers, we search for a feature representation of the state of the hand while not including explicit pose information about the object. Hence, the state of the hand includes measurable features, either tactile or kinesthetic ones, that can be mapped to the true position and orientation of the object. Using such state representation, we study fundamental tasks for in-hand manipulation without relying on visual perception: learning an haptic-based pose observation model of a grasped object and controlling the motion of the hand to manipulate the object to some goal. Furthermore, we aim to understand the data requirements and learning abilities over a set of test objects. While an observation model enables somewhat estimation of the pose of the object, the accuracy of a learned model is limited and reaches saturation with some amount of data. Hence, we train a \textit{critic} model with independent data to estimate prediction errors. The critic is then incorporated in a Model Predictive Control (MPC) \cite{Lee2011} to manipulate an object to a goal position while minimizing estimation errors along the motion. In such way, the controller will be able to avoid regions where the observation model tends to provide large erroneous predictions. The scope of this work is focused on the information requirement for haptic-based object pose estimation. While we do not aim for object-agnostic models, we test the ability of a trained observation model to generalize to new objects of different shapes that were not included in training.  

To summarize, the contributions of this work are as follows:
\begin{itemize}
    \item A novel 3D printed simple and low-cost tactile finger is proposed for underactuated hands.
    
    \item An observation model is proposed to map haptic sensing of an underactuated hand to the pose of an object. The haptics include sensing from the tactile fingers along with kinesthetic sensing from the actuators of the hand. With such, a study is conducted about the haptic information required for object pose estimation with an underactuated hand while including tactile and kinesthetic sensing. 
    
    \item We analyze the partial-occluded pose-estimation problem where an initial visual glance of the grasp is available prior to performing the task in an occluded region.
    
    \item Models for pose estimation are shown to be somewhat generalizable.
    
    \item We propose an MPC approach to manipulate a grasped object to desired goal positions solely based on the predictions of the observation model. While an entire pose estimation is available, controlling the orientation of the object is highly dependent on the initial grasp. Hence, the MPC can only control the position of the grasped object. To include orientation, a designated motion planning with some re-grasping maneuvers is required in future work.
    
    \item A critic model is added to the MPC approach such that the controller can reason about the accuracy of the observation model and avoid potentially erroneous actions.
\end{itemize}
To the best of the author's knowledge, this is the first analysis of pose estimation with underactuated hands solely by using simple and low-cost haptic perception. Achieving haptic-based pose estimation would enable further fusion with vision for better estimation in unstructured or partly-occluded environments.

\section{Related work}
\label{sec:related_work}
% In recent years, The usage of tactile feedback for in-hand manipulations have become very prominent since it affords a direct mechanism for reacting to the environment through contact. A wide range of tactile sensor have been developed for robotic hands, from pressure based sensors \cite{DBLP:journals/corr/abs-2101-05452},\cite{barometer}, to optic based detection \cite{multicurve} and vision based sensors  \cite{DBLP:journals/corr/abs-2104-01167} \cite{DBLP:journals/corr/abs-2005-14679}. Theses touch sensors encode essential information for grasping such as contacts location and forces, which can leverages to perceive properties of the grasped objects. Despite the existence of many different tactile sensors, there is a limited selection of commercial tactile-based hands, where the main barriers are the lack of cheap, robust, easy-to-use and easily-fabricated sensors. Moreover, interpretation of raw tactile data from existing finger requires sophisticated data-driven model, which might not always be available and can regress to estimate valuable features.

The presented work discusses the use of underactuated hands in occluded environments by using haptics. Related work on these topics is briefly discussed below.

% Underactuated hands
\textbf{Underactuated robotic hands.} Underactuated hands are %tendon-based 
compliant mechanisms that exhibit complex behavior due to the passivity of the adaptive joints \cite{Sintov2019}. Compliance can be achieved in various mechanisms such as inflating soft materials \cite{Under1,Under2}, or through passive joints with springs and tendons \cite{Dollar2010}. The work in this paper will focus on tendon-based underactuated hands seen in Figure \ref{fig:FlX-finger}a. The recent introduction of open-sourced hands fabricated through 3D printing has enabled easy access to low-cost hardware \cite{Ma2017YaleOP}. However, precise mechanical properties such as joint stiffness, contact coefficients and joint friction are hard to extract. Consequently, accurate models of such systems are difficult to formulate. Modeling tools, introduced in several works \cite{Odhner2011, Rocchi2016, Hussain2018}, examine joint configurations, torques, and energy with a simplified frictional model. Nevertheless, these techniques have been shown to be sensitive to assumptions in external constraints and are generally suitable for simulations. Recent work has proposed the self-identification of necessary parameters through exploratory hand-object interactions using an external camera and particle filtering \cite{Hangeabe1321}.

\textbf{Manipulation in occluded environments.} Additional sensor modalities are commonly used to augment or alternate vision \cite{Bicchi1988}. Most work utilizes haptic sensors to reason about the state of contact \cite{Funabashi2015,Melnik2021}. Haptic perception is generally used to learn various features of an object in uncertain environments to grasp and manipulate it. Such information may include stiffness, texture, temperature variations and surface modeling \cite{Su2012}. Often, haptic perception is used alongside vision to refine initial pose estimation \cite{Bimbo2013}. %The combination of haptic and visual perception, therefore, copes with the limits of solely using vision in a an occluded environment. 
For instance, recent work has trained a policy to leverage multimodal feedback in contact-rich manipulation tasks from tactile and vision through self-supervision \cite{Lee2019}. Another work used 6-axis force-torque and tactile sensors on a multi-finger rigid hand to train temporal neural-network models for in-hand manipulation tasks \cite{Funabashi2020}. Opposed to contact sensing, a different approach used acoustic perception to gain information regarding hand-object contact \cite{Du2022}. Radio frequency perception was used in a different work to retrieve objects in fully-occluded settings \cite{Boroushaki2021}. 

% In-hand manipulation with/without vision
% \textbf{Haptic perception} is commonly used to learn various features of an object in uncertain environments to grasp and manipulate it. Such information may include stiffness, texture, temperature variations and surface modeling \cite{Su2012}. Often, haptic perception is used alongside vision to refine initial pose estimation \cite{Bimbo2013}. As mentioned before, such approach can be limited in poor lighting and occlusions.

% Contact sensing Pose estimation relative to hand
\textbf{Contact sensing.} The common approach for pose estimation during manipulation is tactile to sense contact. Such sensing has been generally achieved using simple force or pressure sensors \cite{Tegin2005}. As such, Koval et al. \cite{Koval2013} used contact sensors and particle filtering to estimate the pose of an object during contact manipulation. In recent years, sensor arrays have become more common due to advancements in fabrication abilities and due to their effectiveness in covering large contact areas \cite{Bimbo2016}. The work by Sodhi et al. \cite{Sodhi2020}, for instance, used data from a optical tactile sensor to estimate the pose of an object being pushed. Nonetheless, an analysis in \cite{Vazquez2014} has proven that sensor complexity does not necessarily provide better performance mainly due to the way tactile data is represented. It was shown that simpler fingertip sensors yielded equivalent accuracy.

% \textbf{Discuss about}:
% \begin {enumerate}
% \item Complexity of (and) modeling underactuated hand 

% % \cite{Odhner2011} - Dexterous Manipulation with Underactuated Elastic Hands 

% \item Data based methods for learning pose model, transition model, correction model 

% \cite{shaj2020actionconditional} Action-Conditional Recurrent Kalman Networks For Forward and Inverse Dynamics Learning (transition model - general)

% \cite{pmlr-v87-golemo18a} - Sim-to-Real Transfer with Neural-Augmented Robot Simulation (transition model - general) 

% % \cite{Sintov2019} -  Learning a State Transition Model of an Underactuated Adaptive Hand (transition, underactuated) 

% \cite{lutter2020differentiable} - Differentiable Physics Models for Real-world Offline Model-based Reinforcement Learning ( dynamic model, general)

% % \cite{narang2021simtoreal} Sim-to-Real for Robotic Tactile Sensing via Physics-Based Simulation and Learned Latent Projections 

% % \cite{Koval2013} - Pose estimation for contact manipulation with manifold particle filters ( pose with sensors) 

% % \cite{Bimbo2016} - In-Hand Object Pose Estimation Using Covariance-Based Tactile To Geometry Matching 

% % \cite{Sodhi2020} - Learning Tactile Models for Factor Graph-based State Estimation 

% \item Planing with underactuated hand.

% \cite{morgan2021visiondriven} morgan2021visiondriven \\

%  \cite{Lepora2020} - 3D-POSE ESTIMATION FROM TOUCH \\ 

% \item Planing with data driven models.

% \end {enumerate} 

\section{Problem Formulation}
\label{sec:probdef}

We consider a two-fingered adaptive hand comprised of two opposing tendon-based fingers as seen in Figure \ref{fig:FlX-finger}a. Each finger has two compliant joints with springs where a tendon wire runs along its length and is connected to an actuator. Also, each distal link of the finger has high friction pads to avoid slipping. Let $\ve{x}\in\mathcal{C}$ and $\ve{a}\in\mathcal{U}$ be some observable state of the hand and action vector, respectively, where $\mathcal{C}\subset\mathbb{R}^n$ and $\mathcal{U}\subset\mathbb{R}^{2}$. Action $\ve{a}$ is the change of actuator angles (i.e., pulling or releasing the tendons) over a fixed time step $\Delta t$. The true state of the hand is not entirely accessible. Hence, we consider a feature state in space $\mathcal{C}$ that may correspond to different kinesthetic features (e.g., actuator angles and loads) and tactile feedback. Also, the system is governed by a transition function $f:\mathcal{C}\times\mathcal{U}\rightarrow\mathcal{C}$ such that, given the current state $\ve{x}_t$ and action $\ve{a}_t$, the next state is given by $\ve{x}_{t+1}=f(\ve{x}_t,\ve{a}_t)$. %For probabilistic transition models, we consider some  parameterized distribution of the form $f_{\theta}(\ve{x}_{t+1}|\ve{x}_t,\ve{a}_t)=P_{\theta}(\ve{x}_{t+1}|\ve{x}_t,\ve{a}_t)$ , overloading notation.

\begin{figure}[]
  \centering
  \includegraphics[width=\linewidth,keepaspectratio]{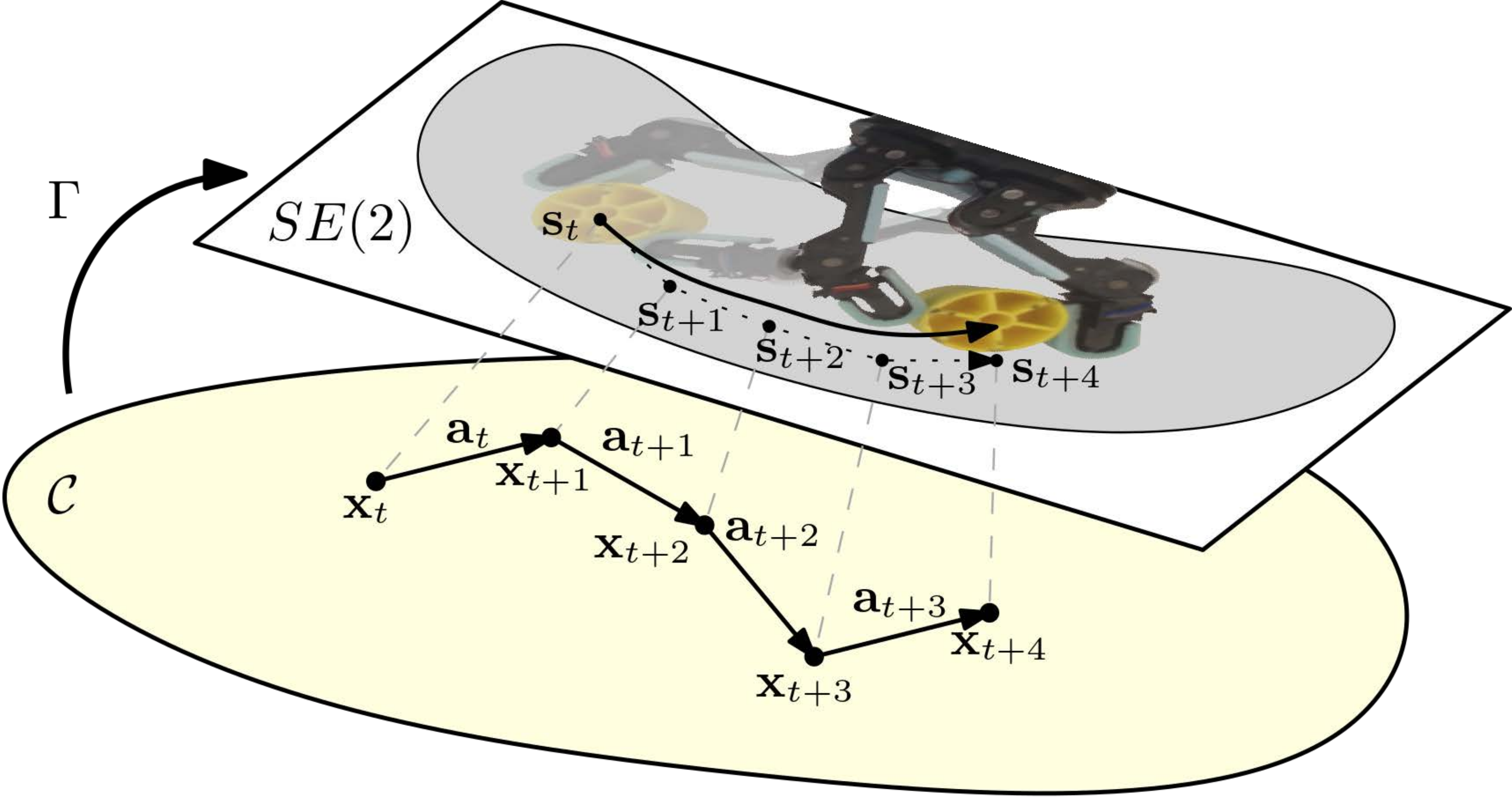}
  \caption{Haptic-based feature space $\mathcal{C}$ where motion is governed by a transition model such that $\ve{x}_{t+1}=f(\ve{x}_t,\ve{a}_t)$. The observation model maps a state $\ve{x}_t$ in $\mathcal{C}$ to the corresponding pose $\ve{s}_t$ of the grasped object in $SE(2)$. Hence, a path from $\ve{x}_t$ to $\ve{x}_{t+4}$ is mapped (gray dashed lines) with the observation model $\Gamma$ to the approximated path of the object in $SE(2)$ (dotted curve from $\ve{s}_t$ to $\ve{s}_{t+4}$). The solid curve in $SE(2)$ denotes the true path of the object. }
  \label{fig:infer}
  \vspace{-0.3cm}
\end{figure}

As stated in the introduction and due to inherent fabrication uncertainties, neither the true state of the hand nor an analytical formulation of the transition model are known. In addition, feedback regarding the true pose of the manipulated object is not always available with vision, for instance, in an occluded environment. The goal is therefore twofold. First, we search for $n$ state features that are easy to measure and best embed the pose $\ve{s}\in SE(2)$ of a grasped object. Such state representation should enable training an observation model $\Gamma:\mathcal{C}\rightarrow SE(2)$ to estimate the pose of the object. Second, we wish to explore the use of such state representation to manipulate objects without external pose feedback. Hence, we utilize a learned state transition model $\tilde{f}$ to plan and control motion in the feature space $\mathcal{C}$ while acquiring desired positional motion in the work-plane. The above formulation is illustrated in Figure \ref{fig:infer}. We note that it may be possible to learn a direct model for current state and action to the next pose, i.e., $h:\mathcal{C}\times\mathcal{U}\rightarrow SE(2)$. However, such model would not enable propagation as required in planning and Reinforcement Learning (RL).

% learn probability distributions over the transition function $f$. Uncertainty over $f$ is due to hidden state variables that play a role in the transition but cannot be easily measured, as well as a limited amount of data. We learn forward transition model by fitting an approximation $\tilde{f}$ of the true transition function of the sensor states $f$, given the measured trajectories. Once a sensors state transition function $\tilde{f}$ and a pose estimator $\tilde{g}$ are learned, we can plan in the $SE(2)$ space using $\tilde{f}$ to predict the distribution over sensor-state trajectories and $\tilde{g}$ to predict the distribution of the pose at each sensor-state. 

%%%%%%%%%%%%%%%%%%%%%%%%%%%%%
%%%%%%%%%%%%%%%%%%%%%%%%%%%%%

\section{Learning an Observation Model}
\label{sec:Observation}

In this section, we discuss the possible kinesthetic and tactile features to be included in the feature state representation. This includes the design and fabrication of the FlX-finger. %, a low-cost tactile finger.
Further, we discuss data-based models to learn model $\Gamma$.

\subsection{FlX-finger: A low-cost tactile finger pad}
% ForceSight
% FlexTouch
% FlexiFinger
% Flexible Force Pad Sensor - FFP, FlxForcePad

We present the design of a low-cost tactile finger termed \textit{FlX-finger}. FlX-finger is a flexible 3D printed mechanism that augments the sensor and provides more contact data. It is mostly fabricated through 3D printing and its shape is based on the design of the distal finger link of the Model-T42 hand \cite{Ma2017YaleOP}. The FlX-finger is composed of four parts seen in Figure \ref{fig:FlX-finger}b: finger body, contact pad, FSR sensor and sensor socket. The finger body is connected to the base of the contact pad solely through a flexible joint. The entire finger body including the base of the contact pad is a one-shot 3D print. After print, we cast Urethane Rubber on the base of the contact pad to acquire an high-friction and soft surface as in \cite{Ma2017YaleOP}. The sensor socket is also printed and contains the FSR. Hence, the sensor socket with the FSR are positioned fixed to the finger body within a designated groove. At such configuration, the FSR is in contact with a small bulge on the back side of the contact pad as seen in Figure \ref{fig:FlX-finger}c. The bulge applies force on the FSR when pressure is exerted on the contact pad and the flexible joint deforms. 

FSR sensors are made of polymer films that vary their electrical resistance upon changing pressure on their surface. They are simple to use and low-cost. The FSR sensor is connected to an analog pin of the Arduino Nano through a voltage divider of 4.7k$\Omega$ resistor. When an object is being pressed against the contact pad, the flexible joint and pad deform and press the bulge onto the FSR. The contact distance from the bulge across the contact pad along with the magnitude of the normal contact force would define the force on the FSR and, therefore, the measured voltage. Hence, different contact locations across the pad provide different voltages. Consequently, pressure on the contact pad provides voltage signals that embed information regarding the location and magnitude of contact. To acquire somewhat repeatability of measurements, leveling screws on the back side of the finger body are used to vary the height of the sensor socket and, thus, tune the initial voltage on the FSR. By standard, the initial voltage $V_o$ is determined with a weight of 100 grams directly placed on the contact pad when in an horizontal posture. Measurements are then taken relative to $V_o$. To sum-up, a standard off-the shelf FSR sensor and a 3D printed finger structure is proposed to acquire diverse contact data when grasping an object. The prototype cost of the FlX-finger is estimated at less than \$10. 

\begin{figure}
\centering
\includegraphics[width=0.95\linewidth,keepaspectratio]{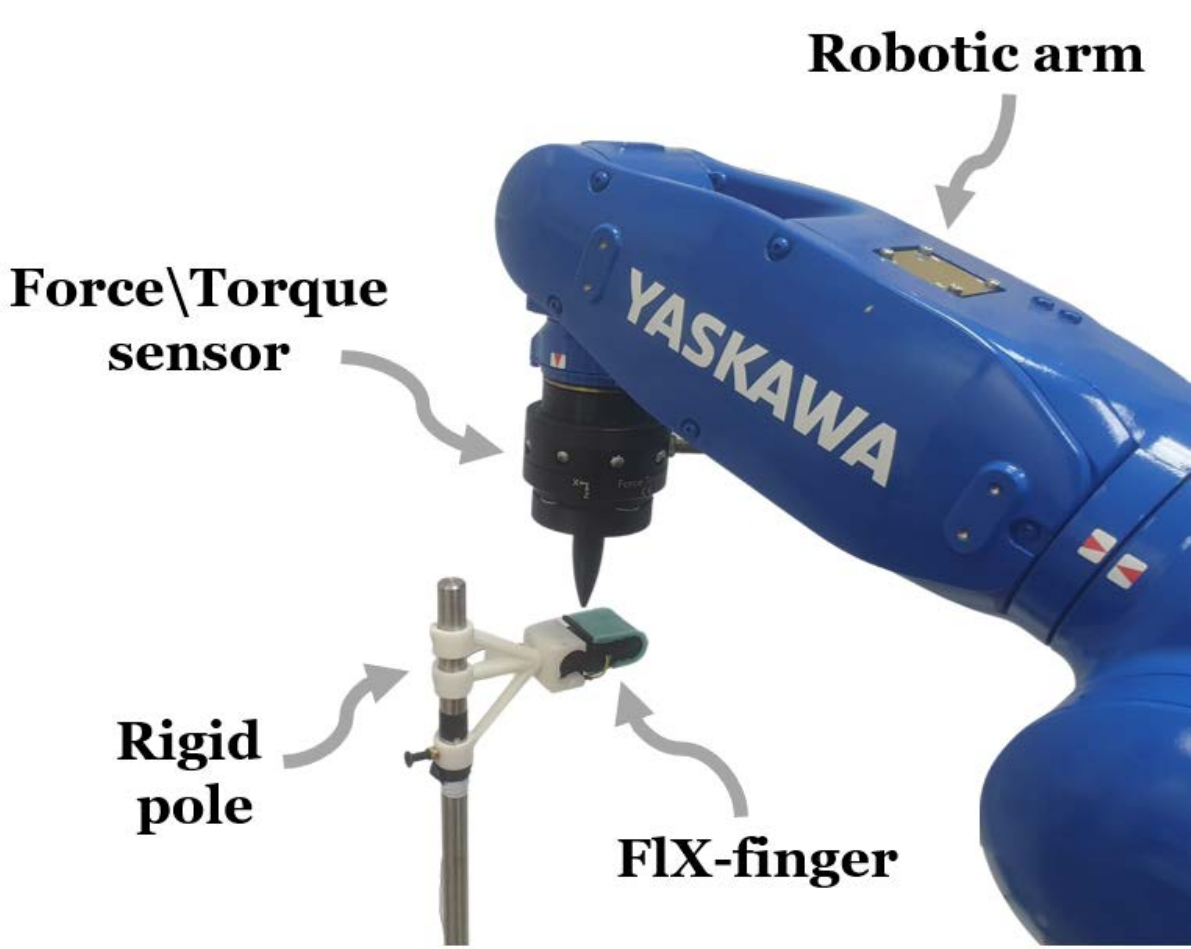}
\caption{Experiment to characterize signal distribution on the FlX-finger.}
\label{fig:FlX-finger_exp1}
\end{figure}

\begin{figure}
\centering
\includegraphics[width=\linewidth,keepaspectratio]{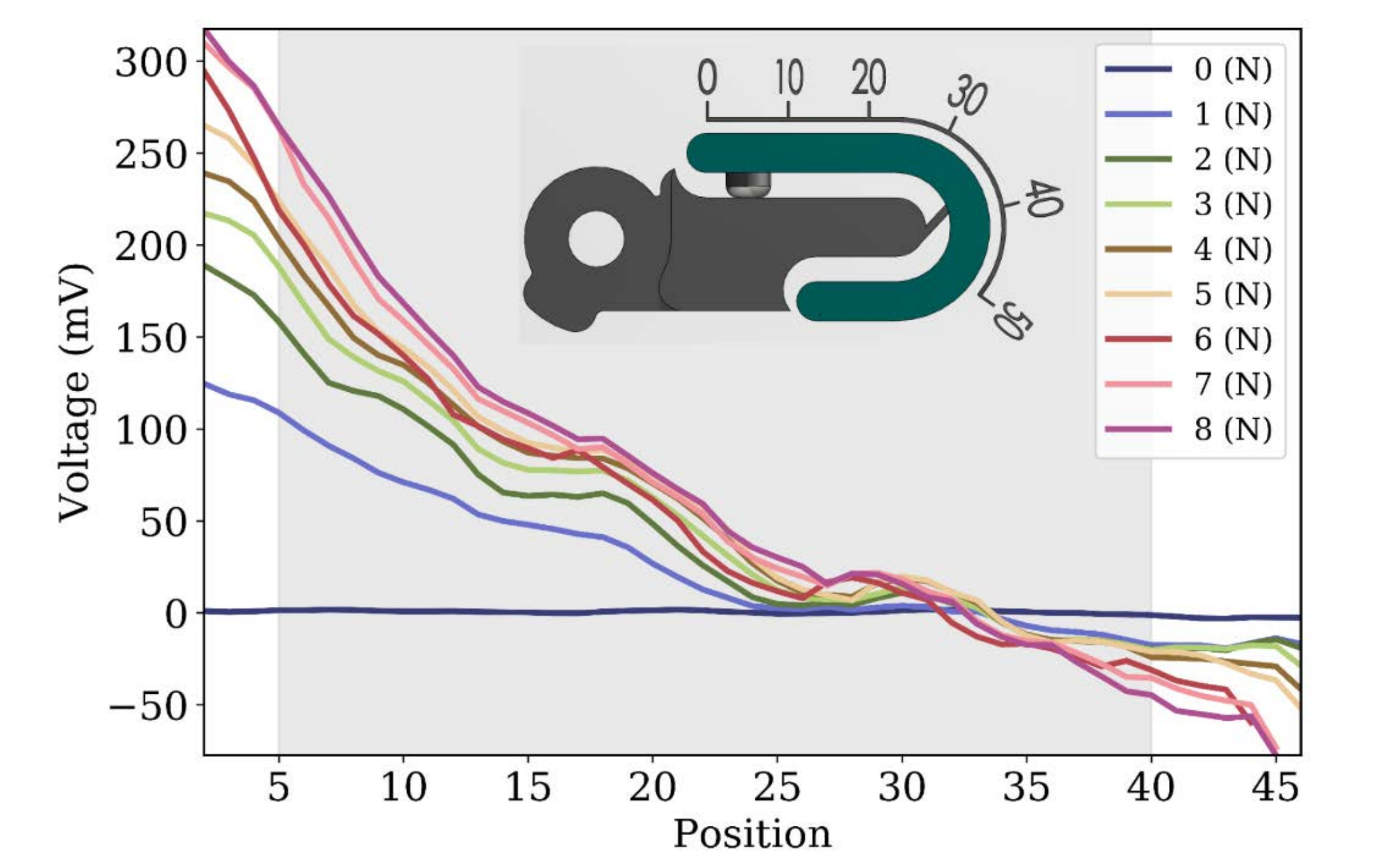} \\
\caption{Signal distribution results for the FlX-finger. The grey area denotes the main working region during manipulation. When the contact is at the back of the finger, force is reduced on the FSR and the signal becomes negative (relative to the initial load).}
\label{fig:FlX-finger_exp2}
\end{figure}

In the formation of the tactile finger, we expect to acquire a force distribution along the contact pad yielding information on contact position. To validate this and acquire a characteristic force distribution of the FlX-finger, we have designed an experiment using a six degrees-of-freedom robotic arm equipped with a Force/Torque sensor and a rigid pole at its tip as seen in Figure \ref{fig:FlX-finger_exp1}. During the experiment, the pole was repeatedly pushed against the finger with predefined forces between $0-10~N$ and at 50 equal length (1~mm) locations along the contact pad. Results for FlX-finger with $V_o= 200~mV$ are seen in Figure \ref{fig:FlX-finger_exp2}. The results show a near linear behaviour in the work region of contact. Due to the spring-like mechanism within the finger, the signal output increases as the contact location approaches the FSR, and decreases when moving away. While the force-distribution of a single FlX-finger cannot be used to identify explicit or accurate contact position, these results indicate that contact information is embedded in the measurements. However, we care about the pose of the object and not explicit contact locations. Hence, we explore the use of two force-distributions from two FlX-fingers along with kinesthetic haptics to build an observation model as described next.

% In addition, results validate the ability to identify the point of contact in an accuracy of approximately 5 mm with various contact loads. 

%%%%%%%%%%%%%%%%%%%%%%%
%%%%%%%%%%%%%%%%%%%%%%%
%%%%%%%%%%%%%%%%%%%%%%%

\subsection{Training data}
\label{sec:training_data}

Training data is collected by grasping an object and manipulating it with random actions. During manipulation, ground-truth data of object poses $\ve{s}_t$ is provided. Thus, the resulting data is a set of observed states, poses and actions  $\mathcal{P}=\{(\ve{x}_1,\ve{s}_1,\ve{a}_1),\ldots,(\ve{x}_N,\ve{s}_N,\ve{a}_N)\}$. In our work, we investigate the contribution of different available features in $\ve{x}$ for learning accurate observation and transition models. These include actuator torques, actuator angles and FlX-finger tactile voltage signals. %Hence, state $\ve{x}$ can be with dimension $n=2$ to $n=6$. 
Given a state definition, we train an observation model $\ve{s}_t=\Gamma_\theta(\ve{x}_t)$ to map a feature state at time $t$ to the true pose of the object. Vector $\theta$ consists of the trained parameters of the model. Model $\Gamma_\theta$ is trained using recorded input and output $\{\ve{x}_i,\ve{s}_i\}_{i=1}^N$ from $\mathcal{P}$. The actions in $\mathcal{P}$ will later be used to train a transition model.

\subsection{Observation Model}

We investigate several architectures of regression models that could approximate $\ve{s}_{t}=\Gamma_{\theta}(\ve{x}_t)$ in real-time.  We test Gaussian Processes (GP) \cite{Rasmussen2005}, Fully-Connected Neural-Network (FC-NN) and Long Short-Term Memory (LSTM) based Recursive-NN (RNN). GP and FC-NN provide prediction for a single time-step, i.e., based on an instantaneous state perception. %With GP regression, the training data is used to acquire a Gaussian distribution about the prediction. Such probability distributions provide a mechanism to quantify model and state uncertainty. 
We employ local GP regression similar to \cite{Sintov2019} where the function’s value at a given query state is predicted using a subset of the training points that are in the vicinity of the query point. %The training data is also used to train and optimize the architecture of a FC-NN. On the other hand, 
LSTM is a time-dependant NN where we exploit internal memory of the motion. Since the force sensors may be slightly noisy, having several consecutive measurements may improve accuracy. Hence, such model is formulated by $\ve{s}_{t}=\Gamma_{\theta}(\ve{x}_t, \ve{x}_{t-1},\ldots,\ve{x}_{t-w})$ where $w$ is the size of past state measurements to include. Observing prediction accuracy of an LSTM model would provide insights about the importance of previous states compared to single state models.
Since the motion of the object is slow while prediction time is very fast (demonstrated in the experimental section), latency between the measured state and prediction is assumed to be negligible. Hence, the predicted pose sufficiently corresponds to the true pose expressed by the feature state.

\subsection{Critic model}
With the addition of data to the training of the observation model, the accuracy improves until reaching saturation at some point. Hence, the accuracy may be limited with the addition of training data. This would be demonstrated in the experiments. Inspired by \cite{icra2020a}, we use redundant training data to generate a critic model $e_t=C_\varphi(\ve{x}_t)$ where $e_t\in\mathbb{R}^+$ and $\varphi$ consists of the trained parameters of the model. Model $C_\varphi$ is trained with measured inputs $\ve{x}_j$ and corresponding labels $e_j=\|\tve{s}_j-\ve{s}_j\|$ where $\tve{s}_j=\Gamma_\theta(\ve{x}_j)$ and $\ve{s}_j$ is the ground-truth pose. The critic provides an estimate of the haptic observation error. With the critic, we would be able to plan and control the system to move in regions with higher observation accuracy as discussed in the next section.

\section{Transition Modeling and Control}
\label{sec:method}

Having a transition model $\ve{x}_{t+1}=f(\ve{x}_t,\ve{a}_t)$ is a key component for planning and control for an underactuated hand. In this section, we discuss the training of a transition model and its usage to control motion. The data described in Section \ref{sec:training_data} is organized in the form $\{(\ve{x}_i,\ve{a}_i),(\ve{x}_{i+1})\}_{i=1}^N$. Unlike previous work \cite{Sintov2019} where actions were from a discrete set with a cardinality of 4-8, actions in our work are continuous within a normalized range $[0,1]$. Similar to previous work \cite{Calli2016, Calli2018a} and since the hand is underactuated, we focus on controlling the position of the object. The ability to reach some orientation is highly dependent on the initial grasp. Hence, including orientation, which is predicted by the proposed observation model, requires a designated motion planning while also considering re-grasping maneuvers (e.g., pick and place). The orientation control problem is left for future work.

As discussed in Sections \ref{sec:introduction}-\ref{sec:related_work}, previous work such as in \cite{Sintov2019} used a vision-based transition model. In another work \cite{Calli2017}, MPC is applied using image-based visual servoing with a linear approximation of the system \cite{Calli2016} without the use of a transition model as in this work. While these methods and others rely on visual perception, we propose an MPC approach that is based on haptics and a data-based model. Hence, an accurate motion to various goals is possible without the need for a line-of-sight with the hand.

% --------------------------------------------------------------

\subsection{Learning a Transition Model}
\label{sec:transition_model}

Given a feature state representation as discussed above, we now learn a transition model to use in the control. A straightforward parameterization of $\tilde{f}_{\phi}(\ve{x}_t,\ve{a}_t)$, where $\phi$ is the a vector of network weights, can be difficult when the sampling frequency is high and the consecutive states are too similar \cite{Nagabandi2018}. Therefore, instead of learning a direct transition function, we train a model to predict the change from state $\ve{x}_t$ given an action $\ve{a}_t$ over time step ${\Delta}t$. The next state prediction is, consequently, given by 
\begin{equation}
    \label{eq:prediction_state}
    \tve{x}_{t+1}=\ve{x}_t + \tilde{f}_{\phi}(\ve{x}_t,\ve{a}_t).
\end{equation}
% We note that the recorded data includes sliding of the object relative to the fingers and, therefore, sliding is modeled.
Several data-based models are benchmarked for training model $\tilde{f}_\phi$ including GP \cite{Sintov2019} and FC-NN. To capture deviations that could violate the standard Markov assumption on the state space, we also learn a time-dependent LSTM-based RNN such that $\tve{x}_{t+1}=\ve{x}_t + \tilde{f}_{\phi}(\ve{x}_t,\ve{x}_{t-1}\ldots,\ve{x}_{t-k},\ve{a}_t)$ where $k>0$ is the number of preceding states. We note that the data used for learning the transition model includes sliding of the object along the contact pads relative to the fingers in some states and actions. Hence, sliding is modeled and taken into account in planning and control. 

% 

% \subsubsection{Transition model}
% A straightforward parameterization of $\tilde{f}_{\phi}(x_{t+1}|x_t,a_t)$ can be difficult when the consecutive states are too similar \cite{DBLP:journals/corr/abs-1708-02596}, hence instead of learning direct transition function, we trained a transition model to predict the change in state $x_t$ given the action $a_t$ over the time step ${\Delta}_t$, i.e: $x_{t+1}=\tilde{f}_{\theta}(x_t,a_t)+x_t$. Experimenting with the system, we heuristically defined a constant time step ${\Delta}_t$ to equal 0.1 seconds, to increase the information of each transition while avoiding decreasing the underlying continuity of the system dynamics. We used the same 1-step time dependant GP and MLPs architectures as described in previous Section. 
% To capture deviations that could violate the standard Markov assumption on the state space, we also learned a time-dependent RNN (implemented as LSTM) $x_{t+1}=\tilde{f}_{\theta}(x_{t:t-1},a_t)+x_t$.

% --------------------------------------------------------------

\subsection{Haptic Servoing}
\label{sec:HS}

% We used as baseline the Visual Servoing (VS) based linear approximation of a two-finger hand which maps the object velocity $v_o$ to the required actuator velocity $\dot{q}$ : 
Previous work by Calli and Dollar \cite{Calli2016} have proposed the use of Visual Servoing (VS) to control the hand for reaching a goal position and tracking a path. VS utilizes a simple control rule based on linear approximation of the hand-object system and is only able to control the position of the object (without orientation). The control maps the desired object velocity to the required actuator velocity and is given by \cite{Calli2016}
\begin{equation}
    \ve{a}_t= 
    \begin{bmatrix} 
    \frac{1}{K_x} & \frac{1}{K_y} \\
    -\frac{1}{K_x} & \frac{1}{K_y}
    \end{bmatrix}
    \ve{v}_t ,
\end{equation}
where $\ve{v}_t=\ve{p}_g-\ve{p}_t$, $\ve{p}_g\in \mathbb{R}^2$ is the desired goal position, $\ve{p}_t$ is the current position extracted from $\ve{s}_t$ and $K_x,K_y>0$ are constant scalars related to the hand. Position $\ve{p}_t$ is originally measured in real-time using visual perception. Let map $\tilde{\Gamma}_\theta:\mathcal{C}\to\mathbb{R}^2$ be a position estimation function taken from the pose estimation map $\Gamma_\theta$. Our proposed Haptic Servoing (HS) approach approximates $\ve{p}_t$ such that $\ve{v}_t=\ve{p}_g-\tilde{\Gamma}_\theta(\ve{x}_t)$. Hence, the system is driven by measuring the haptic state rather than having visual perception.
%%%%%%%%%%%%%%%%%%%%%%%%%%%%%%
% \begin{figure}[t]
%   \centering
%   \includegraphics[width=\linewidth,keepaspectratio]{figures/control_scheme.png}
%   \caption{Scheme of the proposed MPC approach.}
% \label{fig:mpc_scheme}
% \end{figure}
%%%%%%%%%%%%%%%%%%%%%%%%%%%%%%
%%%%%%%%%%%%%%%%%%%%%%%%%%%%%%
\begin{figure}[]
  \centering
  \begin{tabular}{c}
  \includegraphics[width=\columnwidth,keepaspectratio]{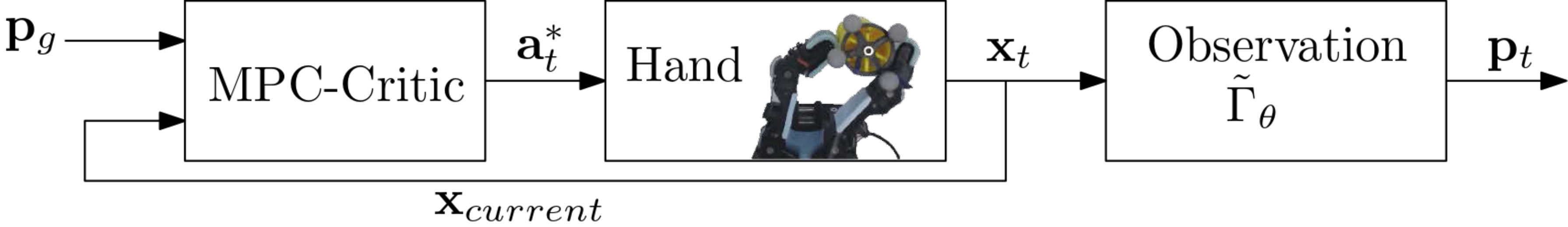}\\\\
        \includegraphics[width=\columnwidth,keepaspectratio]{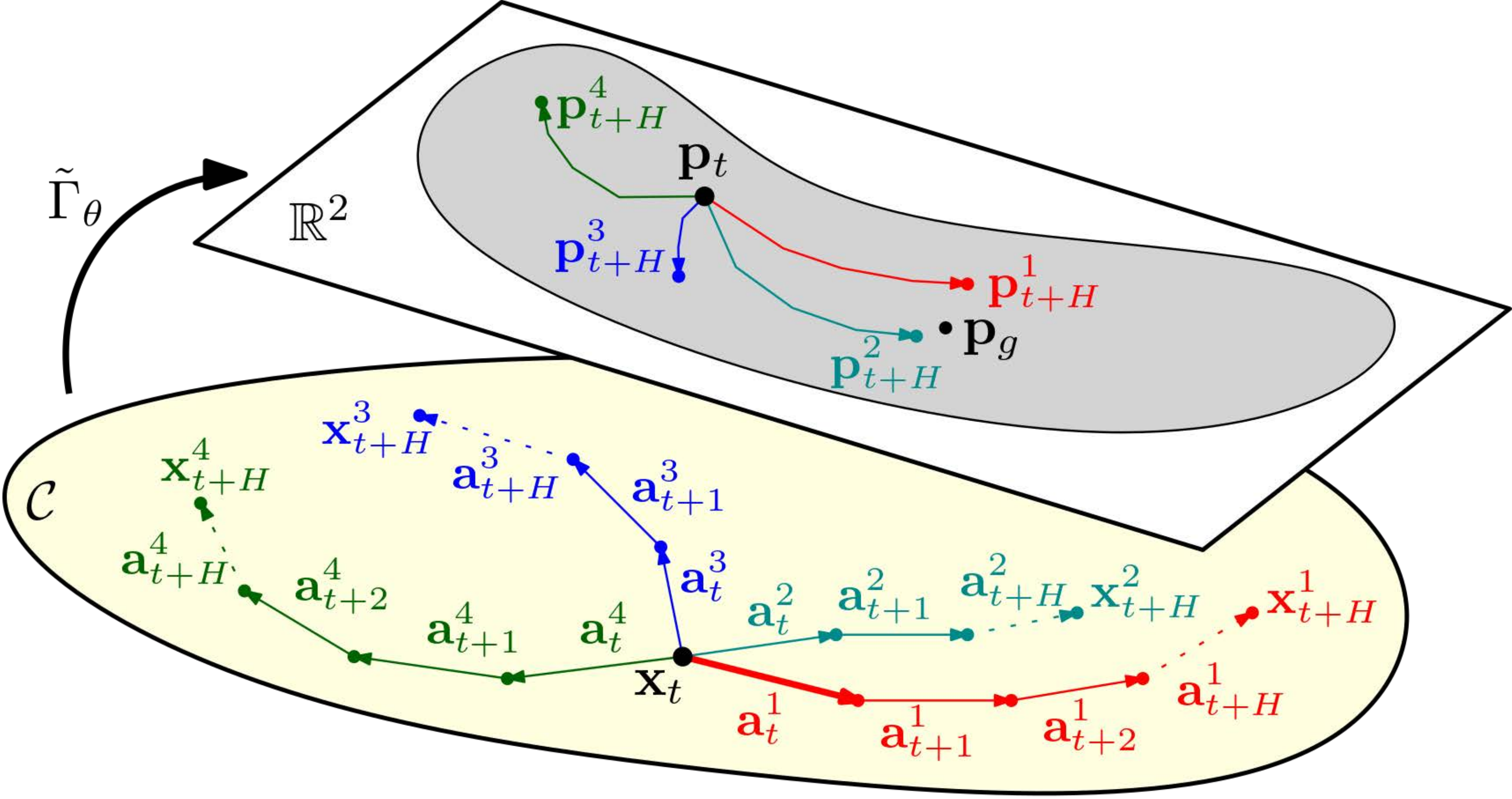}
  \end{tabular}
  \caption{(Top) Control scheme and (bottom) iteration of the shooting method in $\mathcal{C}$ within an MPC control to find the next action to drive the system to goal position $\ve{p}_g$. Random action sequences are applied from the current state $\ve{x}_t=\ve{x}_{current}$ and used to propagate states in $\mathcal{C}$ with the transition model. The resulted paths are inputted to observation model $\Gamma_\theta$ to estimate their positions. The first action of the path that reaches closer to the goal while having a low critic error is chosen. In this example, four sequences are propagated (red, cyan, blue and green). While the cyan path (ending with $\ve{p}^2_{t+H}$) reached closer to the goal than the red one (ending with $\ve{p}^1_{t+H}$), it has higher critic error. Hence, the first action $\ve{a}_t^1$ of the red sequence is exerted on the hand. This process is repeated until reaching the goal position.}
\label{fig:mpc}
\end{figure}
%%%%%%%%%%%%%%%%%%%%%%%%%%%%%%

% In our implementation we directly used VS based on the desired direction of the object toward the intermediate goal point. We also compared VS with the predicted object pose $g_{\theta}(s_t|x_t)$ which we refer as Tactile Servoing (TS).

% --------------------------------------------------------------

% \subsection{Vision-free in-hand manipulations with data-driven models}
\subsection{Haptic-based Planning and Control}
\label{sec:MPC}

Since we have no explicit feedback in $SE(2)$, the above HS control relies solely on pose predictions of the learned observation model. However, the model includes some errors and uncertainties over the state-space. While HS can be useful to reach a near-by position 
% or track a short path 
(as will be demonstrated in the experiments), it does not reason about these possible erroneous predictions of the observation model. Consequently, inaccurate predictions may deviate the motion from the desired goal. Our objective is, therefore, to formulate a controller that can reason about positioning inaccuracies without visual feedback. We propose a Model Predictive Control (MPC) with a random-sampling shooting method. As discussed in \cite{Morgan2020}, MPC can be advantageous for manipulation. In MPC, the next action is optimized based on some cost function while considering the future outcome of candidate paths. The first action of the path with the lowest cost is exerted. 

% Planning without seeing...(Exploding opening sentence) Our objective is the create a planner that computes an optimal sequence of actions for a learned data driven hand which minimize a given cost function. We formulate an online model-based controller utilizing Model Predictive Control (MPC) with random-sampling shooting method. As discussed in \cite{Morgan2020}, MPC is advantageous for manipulation, as the next control input is optimized after each transition, which benefits the uncertainty propagation of the learned models. Once we learned the model, we use it for control by predicting the future outcomes of a candidate trajectories and selects the actions which results in the lowest cost. 
% Given the latent hand state $\ve{x}_t$ at time $t$, the prediction horizon $H$, a random action sequence $A_{t,H} = \{\ve{a}_t,...,\ve{a}_{t+H}\}$, the hand transition model $f_{\theta}(x_{t+1}|x_t,a_t)$, the pose model $g_{\theta}(s_t|x_t)$ and the critic $C_\theta(x_t,g_{\theta}(x_t))$, the MPC controller applies the first of action of optimized action:

Given the current hand state $\ve{x}_t$, prediction horizon $H$ and an action sequence $A_{t,H} = \{\ve{a}_t,...,\ve{a}_{t+H}\}$, the predicted path in $\mathcal{C}$ acquired by propagating model \eqref{eq:prediction_state} for $H$ steps with actions $A_{t,H}$ is $\sigma_t=\{\ve{x}_t,\tve{x}_{t+1},\ldots,\tve{x}_{t+H}\}$. The cost of the path is defined to be \begin{equation}
    \label{eq:cost}
    J(\sigma_t) = \sum_{\ve{x}_\tau\in\sigma_t} \left[ w_1\|\tilde{\Gamma}_\theta(\ve{x}_{\tau})-\ve{p}_g\|^2 + w_2 C_\varphi(\ve{x}_\tau)  \right] 
\end{equation}
where $\ve{p}_g\in \mathbb{R}^2$ is the goal position and $w_1,w_2>0$ are pre-defined weights. The first component in the sum of \eqref{eq:cost} estimates the distance of states in the path to the goal. This component prioritize paths that steers more towards the goal. The second component, on the other hand, evaluates the accuracy of the pose approximated by the observation model. We aim to exert the path that takes the object closer to the goal while having high certainty about its pose along the path. Therefore, at time $t$, we search for the set of actions $A^*_{t,H}$ that is the solution to
\begin{equation}
    \label{eq:opt}
    % \begin{array}{lcl}
        A^*_{t,H} = \argmin_{A_{t,H}}  J(\sigma_t).
    %     % & \textrm{s.t.} & \ve{x}_{\tau+1}=\ve{x}_\tau + \tilde{f}_{\phi}(\ve{x}_\tau,\ve{a}_\tau) \\
    %     % && \ve{s}_\tau =  \Gamma_{\theta}(\ve{x}_\tau) \\
    % \end{array}
    % \vspace{-3pt}
\end{equation}

Finding the optimal solution of \eqref{eq:opt}, i.e., optimal sequence of actions $A^*_{t,H}$, could be slow and not suitable for real-time control. Hence, we search for a near-optimal solution in a random-sampling shooting setting. Algorithm \ref{alg:mpc} presents the \textit{MPC-Critic} control with the shooting method. Figure \ref{fig:mpc} presents the control scheme and illustration of the shooting method within the MPC-Critic control. In each iteration and current state $\ve{x}_t=\ve{x}_{current}$, $m$ random action sequences $\{A_{t,H}^{(1)},\ldots,A_{t,H}^{(m)}\}$ are generated and their costs are evaluated according to \eqref{eq:cost}. The first action of the lowest cost sequence is exerted. The process is repeated until reaching within distance $\epsilon$ to the goal $\ve{p}_g$.

%%%%%%%%%%%%%%%%%%%%%%%%%%%%%%
\begin{algorithm}%[H]
    \caption{MPC-Critic($\ve{p}_g$, $H$, $\epsilon$)  }
    \label{alg:mpc}
    \SetAlgoLined
    % \textbf{Input:} $\ve{s}_g$, $H$, $\epsilon$ \;
    \textbf{Get} current hand state $\ve{x}_{current}$\;
    \Do{$\|\tilde{\Gamma}_\theta(\ve{x}_{current})-\ve{p}_g\| < \epsilon$}{ 
    % \For{ Time $t=0$ to task horizon $T$}{
    % \textbf{Get} current hand state $\ve{x}_t$\;
    $\ve{x}_{t}\gets\ve{x}_{current}$\;
    \textbf{Sample} random $\{A_{t,H}^{(1)},\ldots,A_{t,H}^{(m)}\}$\;
    \For{$i\gets 1$ \textbf{to} $m$}{
        % Initialize horizon cost $J^i$ \;
        $\sigma_t\gets\{\ve{x}_t\}$\;
        \For{$\ve{a}_\tau\in A^{(i)}_{t,H}$}{
            % $\tve{p}_{\tau}\gets\Gamma_\theta(\ve{x}_{t})$ \;
            % Update cost $J^{i}(\ve{s}_{\tau},g)$ \;
            $\ve{x}_t\gets\tilde{f}_{\phi}(\ve{x}_t,\ve{a}_\tau)$\;
            $\sigma_t\gets\sigma_t \cup \{\ve{x}_t\}$\;
            }
        $J_i\gets J(\sigma_t)$ \tcp*{sol. \eqref{eq:cost}}
        }
    $j\gets \argmin(\{J_1,\ldots,J_m\})$\;
    $a_t^*\gets$ first action in $A_{t,H}^{(j)}$\;
    \textbf{Execute} $a_t^*$\;
    \textbf{Get} current hand state $\ve{x}_{current}$\;
    % \If{Reached current goal}{
            % Update goal}
 }
\end{algorithm}

\section{Experiments}
\label{sec:experiments}

% \subsection{Setup}

\begin{figure}[]
    \centering
    \includegraphics[width=0.8\linewidth]{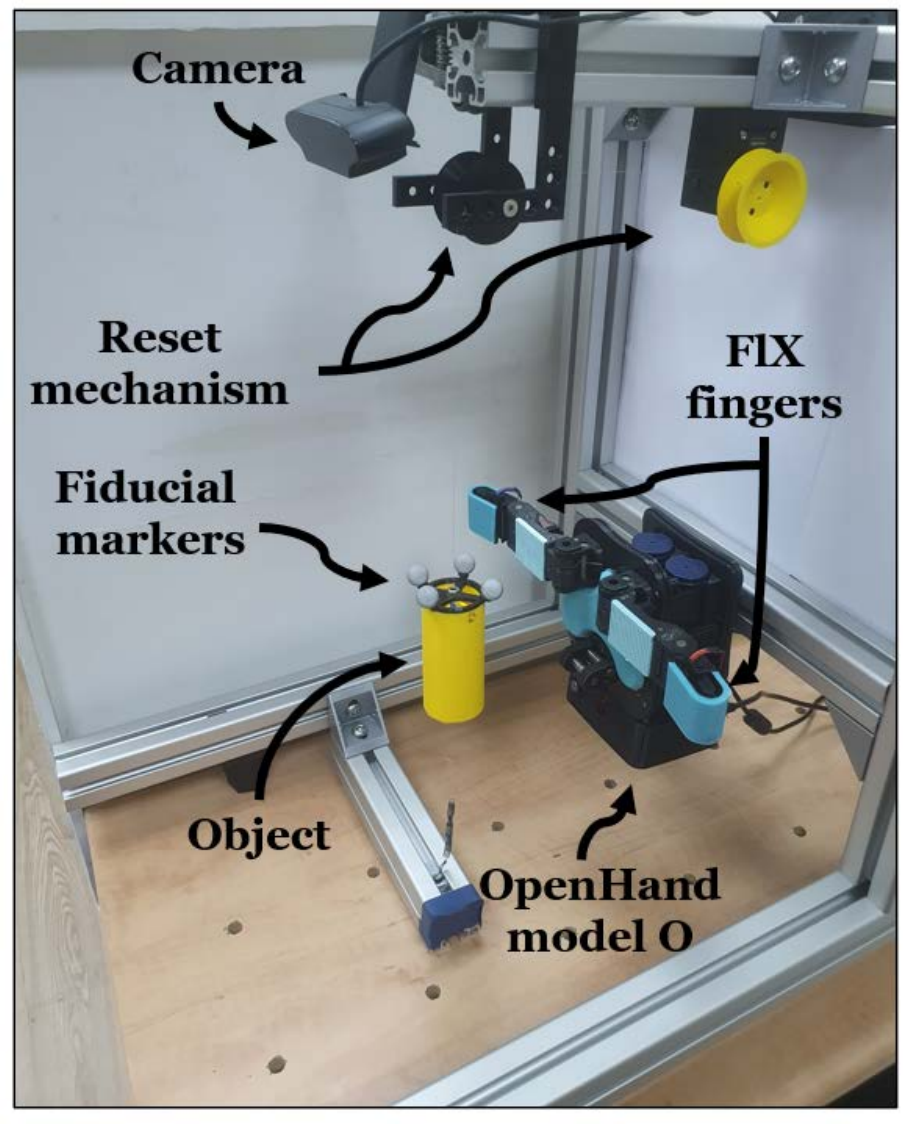}
    \caption{\small Automated experimental setup based on the OpenHand model-O underactuated hand.}
    \label{fig:system}
    % \vspace{-0.5cm}
\end{figure}

\begin{figure}[]
\centering
  \includegraphics[width=\linewidth]{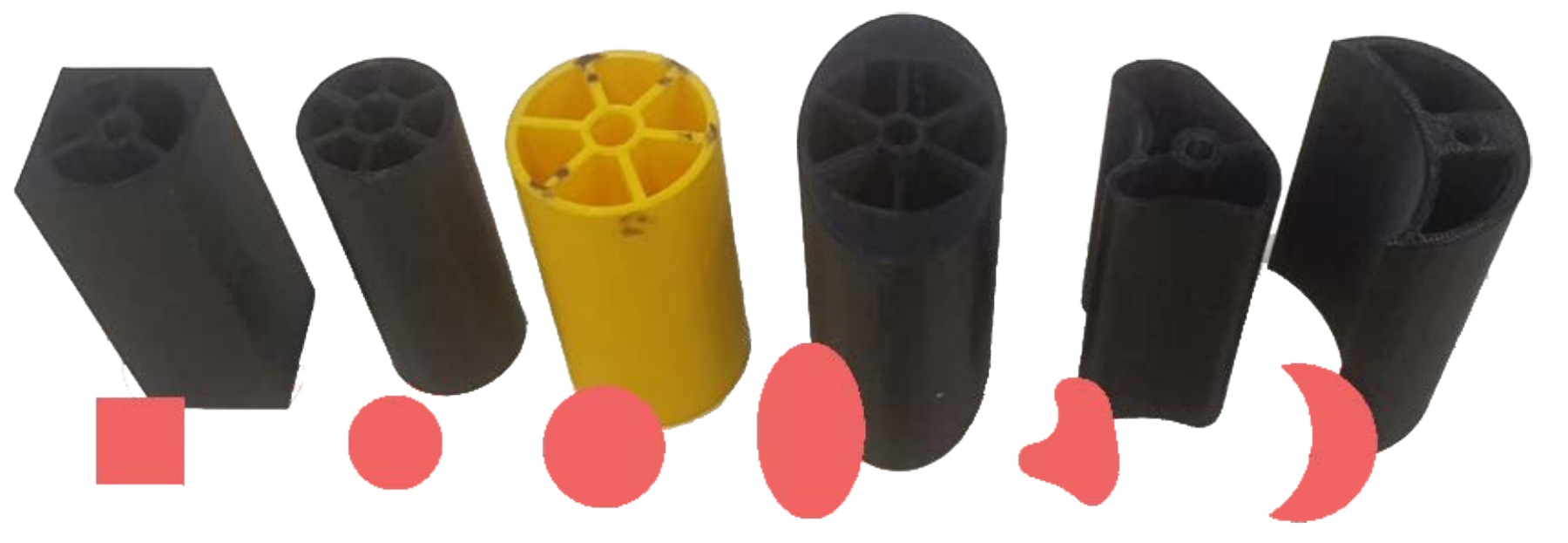} 
  \caption{Six prismatic test objects.}
\label{fig:objects}
\vspace{-0.7cm}
\end{figure}

To validate the proposed approach and analyze the accuracy of various state formulations, we have designed and built an experimental system. We use only the two opposing fingers of the three-finger OpenHand Model-O underactuated hand \cite{Ma2017YaleOP}. The experimental system consists of the hand and an automated reset mechanism as seen in Figure \ref{fig:system}. Hence, training data $\mathcal{P}$ is collected in episodes where, at each episode, the robot grasps the object, performs in-hand manipulation with random actions until it drops, and then repeats. Once dropped, a thin string that runs through a hole in the center of the object pulls the object into the reach of the fingers toward a new grasp. The system is operated by the Robot Operating System (ROS). During manipulation, data stream of various features is available in $10$ Hz and recorded including instantaneous actions, actuator angles and torques, and tactile signals from the FlX-finger. In addition, the pose of the object relative to the hand base is recorded using fiducial markers and cameras. We test eight prismatic objects. Six PLA objects with different cross-sections seen in Figure \ref{fig:objects}: circular (15 mm and 10 mm radius), elliptical,  square, crescent and arbitrary curved. Two additional cylinders of 15~mm radius were used: one flexible (printed with Thermoplastic Polyurethane) and one wrapped with sandpaper for higher friction. Open-source FlX-finger model, fabrication instructions, datasets, open-source code and CAD models of the test objects are available online%in a dedicated Git repository
\footnote{\texttt{https://github.com/osheraz/haptic\textunderscore pose\textunderscore estimation}}.
% \begin{wrapfigure}{r}{0.5\linewidth}
% \centering
%     \begin{minipage}[b]{\linewidth}
%         \includegraphics[width=\linewidth]{figures/system.png}
%         % \vspace{-0.5cm}
%         \caption{\small Automated experimental setup based on the OpenHand model-O underactuated hand.}
%         \label{fig:system}
%     \end{minipage}
%     \begin{minipage}[b]{\linewidth}
%         % \vspace{0.3cm}
%         \includegraphics[width=\linewidth]{figures/objects.png} 
%         % \vspace{-0.7cm}
%         \caption{Six prismatic test objects.}
%         \label{fig:objects}
%     \end{minipage}
%     % \vspace{-0.6cm}
% \end{wrapfigure}
%%%%%%%%%%%%%%%%%%%%%%%%%%%%%%
% \begin{wrapfigure}{r}{0.5\linewidth}
% \end{wrapfigure}
% %%%%%%%%%%%%%%%%%%%%%%%%%%%%%%
% %%%%%%%%%%%%%%%%%%%%%%%%%%%%%%
% \begin{wrapfigure}{r}{\linewidth}
% \end{wrapfigure}
%%%%%%%%%%%%%%%%%%%%%%%%%%%%%%

% ----------------------------------------------------------------
\vspace{-0.3cm}
\subsection{Pose estimation}
We begin by observing pose estimation accuracy for the cylinder of radius 15 mm. The collected training data comprises of $N=92,700$ points recorded by randomly applying actions to the hand in multiple episodes. An additional $10,000$ test points were collected in separated episodes and were not included in the training set in any way. Hyper-parameters optimization of the three tested observation models yielded the following architectures. The FC-NN is composed of 3 layers with 300 neurons and a ReLU activation each. A dropout of $30\%$, early stopping and an L2 regularizer with factor $10^{-5}$ were included to reduce overfitting. For the LSTM based RNN, the optimization yielded a network of 2 layers, 256 hidden neurons each and a sequence of $w=5$ past states. Next, the output is fed into a 256 neurons FC layer with ReLU activation. For both NN models, we used the Adam optimizer along with the MSE loss function. GP with a radial basis kernel used 100 nearest neighbors for local regression.

%%%%%%%%%%%%%%%%%%%%%%%%%%%%%%
\begin{table*}[t!]
\caption{\small Pose estimation accuracy for a cylinder (15~mm) with various feature combinations and regression methods }
\label{tb:pose_tab}
\vspace{-0.3cm}
\begin{adjustbox}{width=1.0\textwidth,center=\textwidth}
\begin{tabular}{c cccc cc cc cc}
% \hline
\toprule
% \multirow{3}{*}{Comb.} & \multicolumn{4}{c|}{Features} & \multicolumn{6}{c|}{Observation Models} \\ \cline{2-11} 
%  & \multirow{2}{*}{\begin{tabular}[c]{@{}c@{}}Actuator \\ angles\end{tabular}} & \multirow{2}{*}{\begin{tabular}[c]{@{}c@{}}Actuator \\ loads\end{tabular}} & \multirow{2}{*}{\begin{tabular}[c]{@{}c@{}}Finger \\ forces\end{tabular}} & \multirow{2}{*}{\begin{tabular}[c]{@{}c@{}}Initial \\ pose\end{tabular}} & \multicolumn{2}{c|}{GP} & \multicolumn{2}{c|}{FC-NN} & \multicolumn{2}{c|}{LSTM $(w=5)$}\\\cline{6-11} 
\multirow{3}{*}{Comb.} & \multicolumn{4}{c}{Features} & \multicolumn{2}{c}{GP} & \multicolumn{2}{c}{FC-NN} & \multicolumn{2}{c}{LSTM}\\%\cline{6-11}
\cmidrule[0.4pt](r{0.125em}){2-5}%
\cmidrule[0.4pt](lr{0.125em}){6-7}%
\cmidrule[0.4pt](lr{0.125em}){8-9}%
\cmidrule[0.4pt](lr{0.125em}){10-11}%
% \cmidrule[0.4pt](lr{0.125em}){8-9}%
% \cmidrule[0.4pt](lr{0.125em}){10-11}%
% \cmidrule[0.4pt](l{0.25em}){12-13}%
& Actuator & Actuator & FlX & Initial & Position & Orientation & Position & Orientation & Position & Orientation \\
& angles & loads & fingers & pose & RMSE (mm) & RMSE (deg) & RMSE (mm) & RMSE (deg) & RMSE (mm) & RMSE (deg)\\\hline
%  &  &  &  &  & \begin{tabular}[c]{@{}c@{}}Position \\ RMSE (mm)\end{tabular} & \begin{tabular}[c]{@{}c@{}}Orientation \\ RMSE (deg)\end{tabular} & \begin{tabular}[c]{@{}c@{}}Position \\ RMSE (mm)\end{tabular} & \begin{tabular}[c]{@{}c@{}}Orientation \\ RMSE (deg)\end{tabular} & \begin{tabular}[c]{@{}c@{}}Position \\ RMSE (mm)\end{tabular} & \begin{tabular}[c]{@{}c@{}}Orientation \\ RMSE (deg)\end{tabular} \\ \hline
1 &  & $\surd$ &  &  & 22.1 $\pm$ 1.7 & 20.1 $\pm$ 1.8 & 20.6 $\pm$ 2.8 & 19.8 $\pm$ 2.7 & 18.8 $\pm$ 2.3 & 18.1 $\pm$ 2.4 \\
2 &  &  & $\surd$ &  & 10.5 $\pm$ 1.4 & 8.7 $\pm$ 1.2 & 10.6 $\pm$ 1.4 & 6.8 $\pm$ 1.9 & 10.1 $\pm$ 1.3 & 6.8 $\pm$ 1.8 \\
3 &  & $\surd$ & $\surd$ &  & 8.5 $\pm$ 1.0 & 8.0 $\pm$ 1.1 & 8.6 $\pm$ 1.2 & 5.9 $\pm$ 0.8 & 7.7 $\pm$ 1.6 & 6.1 $\pm$ 0.8 \\
4 & $\surd$ &  &  &  & 6.6 $\pm$ 1.5 & 8.8 $\pm$ 3.4 & 6.5 $\pm$ 1.2 & 6.1 $\pm$ 1.4 & 5.8 $\pm$ 0.6 & 6.0 $\pm$ 0.8 \\
5 & $\surd$ & $\surd$ &  &  & 6.0 $\pm$ 0.5 & 7.3 $\pm$ 1.8 & 5.7 $\pm$ 1.0 & 6.0 $\pm$ 1.3 & 5.3 $\pm$ 0.6 & 5.5 $\pm$ 1.0 \\
6 & $\surd$ &  & $\surd$ &  & 5.1 $\pm$ 0.6 & 7.0 $\pm$ 1.0 & 4.4 $\pm$ 0.5 & 5.2 $\pm$ 0.6 & 5.1 $\pm$ 0.8 & 4.9 $\pm$ 0.8 \\
7 & $\surd$ & $\surd$ & $\surd$ &  & 4.6 $\pm$ 1.2 & 5.5 $\pm$ 1.8 & 4.3 $\pm$ 0.4 & 5.1 $\pm$ 0.8 & 4.0 $\pm$ 0.2 & 5.0 $\pm$ 0.5 \\\hline
8 & $\surd$ & $\surd$ &  & $\surd$ & 4.3 $\pm$ 1.0 & 5.3 $\pm$ 2.0 & 4.3 $\pm$ 0.7 & 5.0 $\pm$ 0.9 & 4.2 $\pm$ 0.8 & 4.8 $\pm$ 1.0 \\
9 & $\surd$ & $\surd$ & $\surd$ & $\surd$ & 4.1$\pm$ 1.0 & 4.9 $\pm$ 1.7 & 3.9 $\pm$ 0.6 & 4.4 $\pm$ 0.8 & 3.1 $\pm$ 0.6 & 3.0 $\pm$ 0.6 \\ \hline
\end{tabular}
\end{adjustbox}
\vspace{-0.2cm}
\end{table*}
%%%%%%%%%%%%%%%%%%%%%%%%%%%%%%

\begin{figure}[]
  \includegraphics[width=\columnwidth,keepaspectratio]{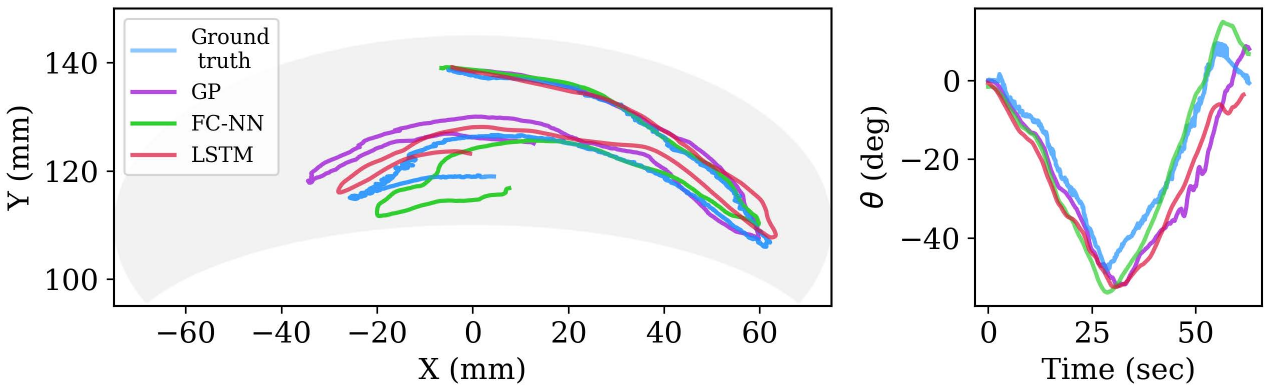}
  \vspace{-0.7cm}
  \caption{ An example of the (left) position and (right) orientation estimation with three observation models over a test trajectory of the cylindrical object. The gray shading illustrates the workspace of the hand approximated with the collected data.}
\label{fig:traj}
\vspace{-0.4cm}
\end{figure}

In our experiments, nine possible feature combinations for a state representation are analyzed. The combinations are listed in Table \ref{tb:pose_tab}. For comparison, we have also included two feature combinations where the initial pose of the object at the moment of grasp is known. Such scenario can occur when a camera has an initial view prior to manipulating the hand in a confined space. In addition, Table \ref{tb:pose_tab} presents accuracy results for the object pose estimation during in-hand manipulation. The accuracy for translation and orientation is the RMSE between predicted and ground truth ones. Figure \ref{fig:traj} shows a trajectory example of a grasped object during in-hand manipulation and the predicted poses along it. Actuator angles clearly provide meaningful information that is essential for accurate estimation. Moreover, the results show that the angles alone can provide relatively good accuracy. Actuator loads provide somewhat accuracy improvement for GP and FC-NN. However, including also tactile information from the FlX-finger exhibits better accuracy for all models. While the FlX-fingers cannot provide accurate predictions by their own (Comb. 2), the results indicate that accuracy is limited by the resolution exhibited in Figure \ref{fig:FlX-finger}c. The results also show that including sequential past states with LSTM improves accuracy compared to single state models (GP and FC-NN) over all feature combinations. Also, having a glance at the ground truth of the initial pose (Comb. 9), if available, exhibits accuracy improvement. The average velocity of the object is 0.62 mm/sec and states are sampled at 10Hz while the LSTM model, for instance, provides a prediction in 9 milliseconds. Hence, an instantaneous sampled state sufficiently corresponds to the predicted pose.

Table \ref{tb:pose_tab_otherObjs} presents additional accuracy results for the remaining objects. The results show that the position estimation exhibits similar or slightly higher errors than the 15~mm cylinder. Results were fairly similar between the flexible and rigid cylinders. Additionally, sliding was not visually observed for the high friction cylinder and, therefore, the model was able to produce more accurate estimations. Furthermore, orientation is shown to be harder to predict as the complexity of the object increases (e.g., crescent and arbitrary objects) and the available haptics do not provide sufficient information. It is assumed that higher resolution tactile sensors such as in \cite{Donlon2018,Lepora2020} would better extract shape features and yield lower orientation errors. Nevertheless and for all objects, having an initial glance at the grasp pose enables accurate orientation prediction.

%%%%%%%%%%%%%%%%%%%%%%%%%%%%%%
%%%%%%%%%%%%%%%%%%%%%%%%%%%%%%
%%%%%%%%%%%%%%%%%%%%%%%%%%%%%%
\begin{figure}[]
  \includegraphics[width=\columnwidth,keepaspectratio]{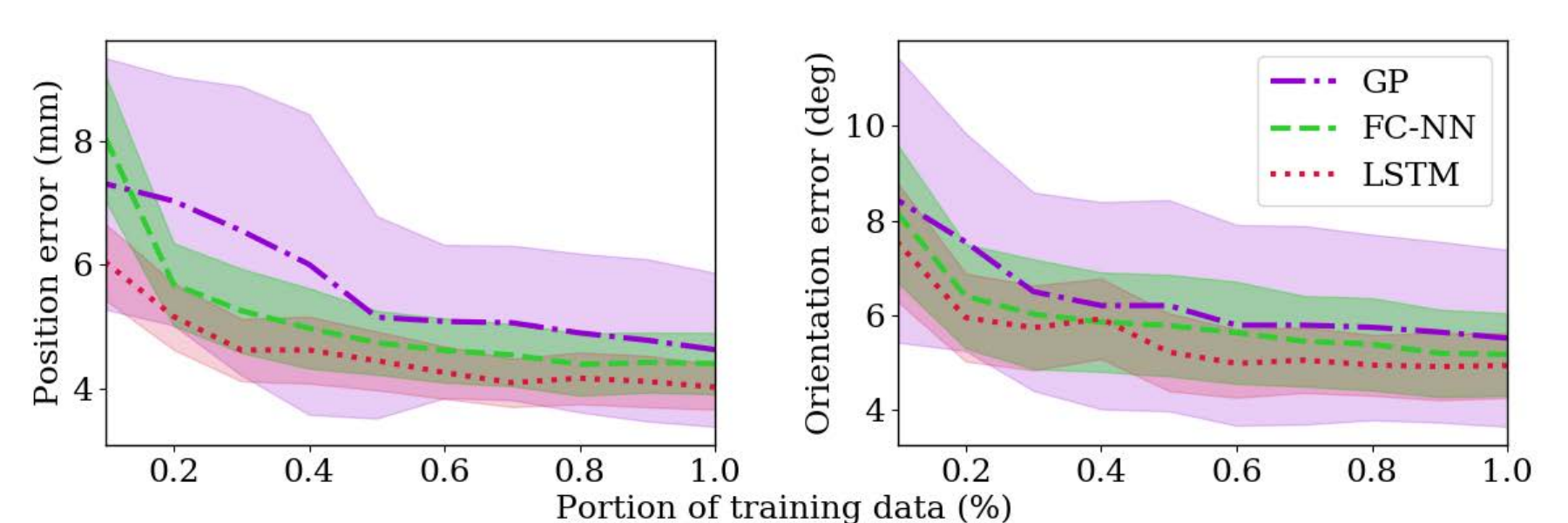}
  \vspace{-0.5cm}
  \caption{Pose estimation accuracy with regards to the percentage of total training data used with GP, FC-NN and LSTM-based RNN.}
  \vspace{-0.3cm}
  \label{fig:portion}
  \vspace{-0.1cm}
\end{figure}
%%%%%%%%%%%%%%%%%%%%%%%%%%%%%%
%%%%%%%%%%%%%%%%%%%%%%%
\begin{table*}[]
\centering
\caption{\small Pose estimation accuracy of the observation model for various objects}
\label{tb:pose_tab_otherObjs}
\vspace{-0.1cm}
\begin{tabular}{c c cc cc cc}
% \hline
\toprule
\multirow{2}{*}{Object cross-section} & \multirow{2}{*}{Comb.} & \multicolumn{2}{c}{GP} & \multicolumn{2}{c}{FC-NN} & \multicolumn{2}{c}{LSTM} \\ %\cline{3-8} 
\cmidrule[0.1pt](r{0.9em}){3-4}%
\cmidrule[0.1pt](r{0.9em}){5-6}%
\cmidrule[0.1pt](r{0.9em}){7-8}%
 &  & \begin{tabular}[c]{@{}c@{}}Position \\ RMSE  (mm)\end{tabular} & \begin{tabular}[c]{@{}c@{}}Orientation\\  RMSE (deg)\end{tabular} & \begin{tabular}[c]{@{}c@{}}Position \\ RMSE  (mm)\end{tabular} & \begin{tabular}[c]{@{}c@{}}Orientation\\  RMSE (deg)\end{tabular} & \begin{tabular}[c]{@{}c@{}}Position \\ RMSE  (mm)\end{tabular} & \begin{tabular}[c]{@{}c@{}}Orientation\\  RMSE (deg)\end{tabular} \\ \hline
\multirow{2}{*}{\begin{tabular}[c]{@{}c@{}}Circular \\ ($r = 10~mm$)\end{tabular}} & 7 & 5.1 $\pm$ 2.1 & 7.0 $\pm$ 4.6 & 4.6 $\pm$ 0.5 & 7.2 $\pm$ 1.1 & 4.4 $\pm$ 0.5 & 7.1 $\pm$ 1.1 \\
 & 9 & 4.8 $\pm$ 2.0 & 6.5 $\pm$ 3.2 & 4.0 $\pm$ 0.8 & 6.0 $\pm$ 2.0 & 4.0 $\pm$ 0.8 & 5.3 $\pm$ 2.0 \\ \hline
 \multirow{2}{*}{\begin{tabular}[c]{@{}c@{}}Circular\\ ($r = 15~mm$, Flexible)~\end{tabular}} & 7 & 5.5 $\pm$ 2.1 & 5.8 $\pm$ 2.8 & 4.7 $\pm$ 0.5 & 4.1 $\pm$ 0.6 & 4.6 $\pm$ 0.4 & 4.0 $\pm$ 0.5\\
& 9 & 3.9 $\pm$ 2.4 & 5.1 $\pm$ 1.7 &  3.9 $\pm$ 0.4 &  4.7 $\pm$ 0.6 &  3.6 $\pm$ 0.3 &  3.9 $\pm$ 0.5\\ \hline
\multirow{2}{*}{\begin{tabular}[c]{@{}c@{}}Circular\\ ($r = 15~mm$, High friction)~\end{tabular}} & 7 & 4.0 $\pm$ 1.1 & 5.5 $\pm$ 2.2 & 4.1$\pm$ 0.4 & 5.3 $\pm$ 0.7 & 3.9 $\pm$ 0.4 & 5.0 $\pm$ 0.6\\
& 9 & 3.5 $\pm$ 1.0 & 4.4 $\pm$ 1.5 &  3.2 $\pm$ 0.4 &  4.9 $\pm$ 0.6 &  2.9 $\pm$ 0.5 &  3.9 $\pm$ 0.4\\ \hline
\multirow{2}{*}{\begin{tabular}[c]{@{}c@{}}Square\\ ( $20\times20~mm$ )\end{tabular}} & 7 & 5.6 $\pm$ 2.4 & 10.7 $\pm$ 6.7 & 5.0 $\pm$ 0.6 & 11.6 $\pm$ 1.3 & 4.6 $\pm$ 0.6 & 11.3 $\pm$ 2.0 \\
 & 9 & 5.1 $\pm$ 2.0 & 6.7 $\pm$ 4.8 & 4.6 $\pm$ 0.6 & 7.5 $\pm$ 1.6 & 4.4 $\pm$ 0.6 & 6.1 $\pm$ 1.0 \\ \hline
\multirow{2}{*}{\begin{tabular}[c]{@{}c@{}}Elliptical\\ ($r_1 = 40~mm,~r_2 = 20~mm$)\end{tabular}} & 7 & 7.5 $\pm$ 2.0 & 30.1 $\pm$ 11.2 & 7.4 $\pm$ 1.1 & 26.1 $\pm$ 1.4 & 6.2 $\pm$ 0.7 & 17.4 $\pm$ 3.6\\
& 9 & 3.8 $\pm$ 1.7 &  5.7 $\pm$ 1.8 &  4.3 $\pm$ 0.5 &  6.0 $\pm$ 0.5 &  4.0 $\pm$ 0.7 &  5.6 $\pm$ 0.6\\ \hline
\multirow{2}{*}{\begin{tabular}[c]{@{}c@{}} Crescent\\ ($r_{out} = 50~mm,~r_{in} = 40~mm$)\end{tabular}} & 7 & 6.2 $\pm$ 4.3 & 27.3 $\pm$ 7.1 & 7.6 $\pm$ 1.5 & 28.2 $\pm$ 5.1 & 7.2 $\pm$ 0.5 & 21.6 $\pm$ 4.0\\
& 9 & 5.8 $\pm$ 4.0 &  6.8 $\pm$ 6.2 &  5.1 $\pm$ 1.0 &  9.6 $\pm$ 1.6&  4.5 $\pm$ 0.8&  7.3 $\pm$ 1.8\\ \hline
\multirow{2}{*}{\begin{tabular}[c]{@{}c@{}}Arbitrary\\ ~\end{tabular}} & 7 & 5.9 $\pm$ 2.6 & 33.1 $\pm$ 4.1 & 6.5 $\pm$ 0.5 & 39.2 $\pm$ 6.1 & 5.0 $\pm$ 0.2 & 28.6 $\pm$ 3.4\\
& 9 & 5.8 $\pm$ 2.5 & 9.0 $\pm$ 6.6 &  5.5 $\pm$ 0.6 &  11.2 $\pm$ 1.9&  4.4 $\pm$ 0.4 &  7.7 $\pm$ 3.0\\ \hline
\end{tabular}
\vspace{-0.3cm}
\end{table*}
%%%%%%%%%%%%%%%%%%%%%%%

%%%%%%%%%%%%%%%%%%%%%%%%%%%%%%
\begin{figure*}
\centering
\begin{tabular}{cc}
\includegraphics[width=0.5\linewidth]{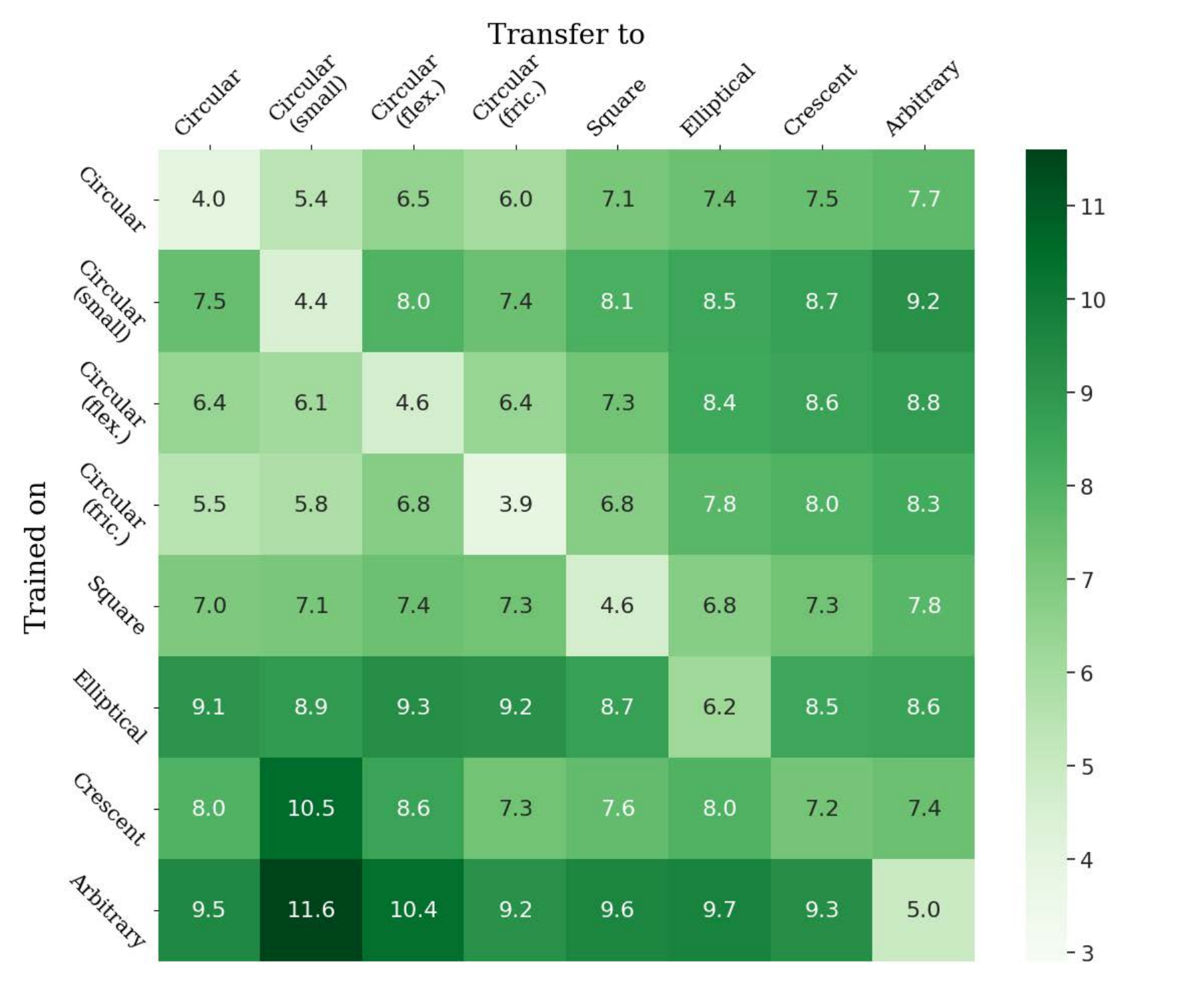} &
\includegraphics[width=0.5\linewidth]{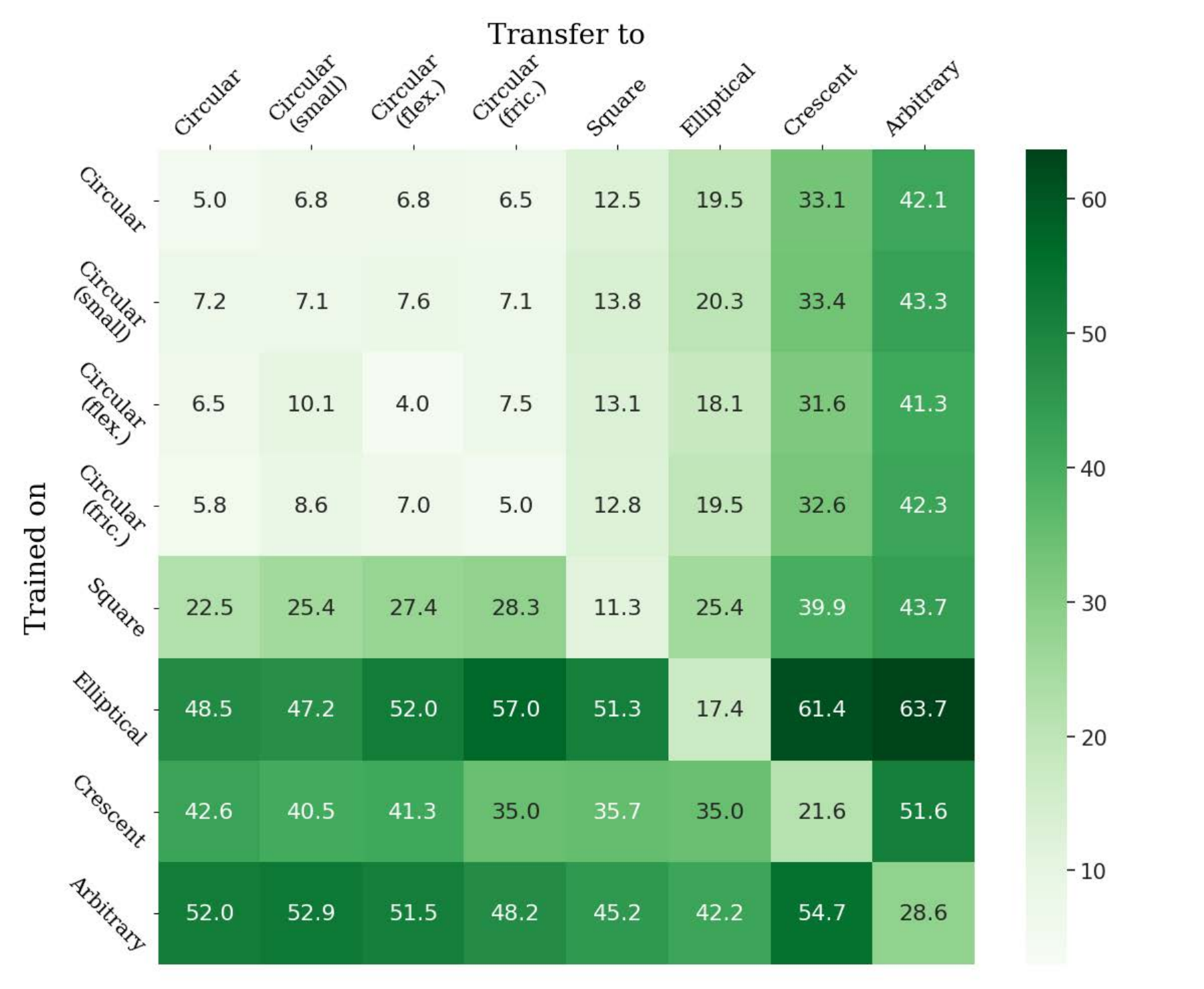} \\ 
\end{tabular}
\caption{Cross transfer pose estimation accuracy between the tested objects with feature combination 7. Transfer scores include (left) position RMSE in \textit{mm} and (right) orientation errors in \textit{degrees}.}
\label{fig:CMconf7}
\end{figure*}
%%%%%%%%%%%%%%%%%%%%%%%%%%%%%%
%%%%%%%%%%%%%%%%%%%%%%%%%%%%%%
\begin{figure*}
\centering
\begin{tabular}{cc}
\includegraphics[width=0.5\linewidth]{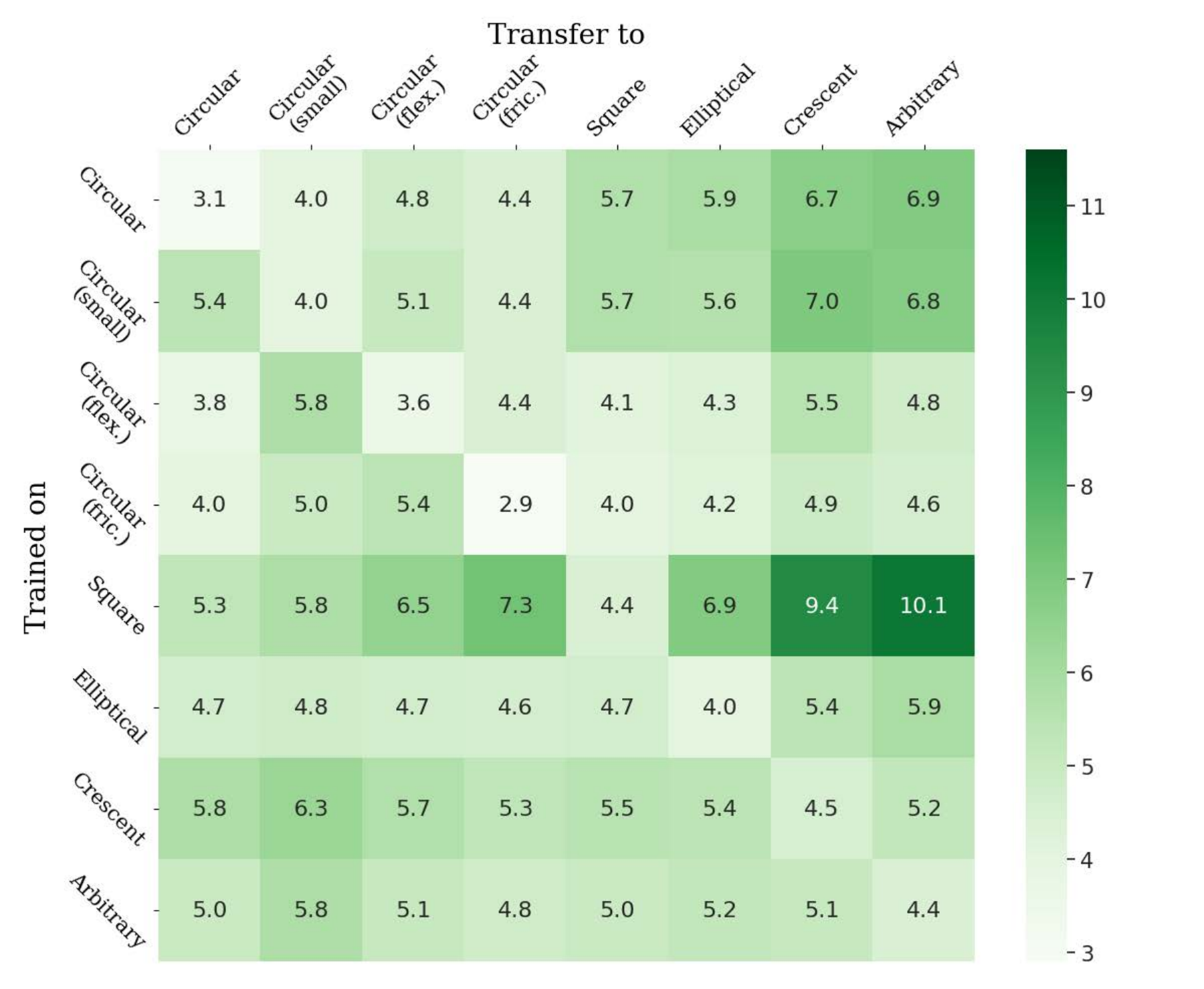} &
\includegraphics[width=0.5\linewidth]{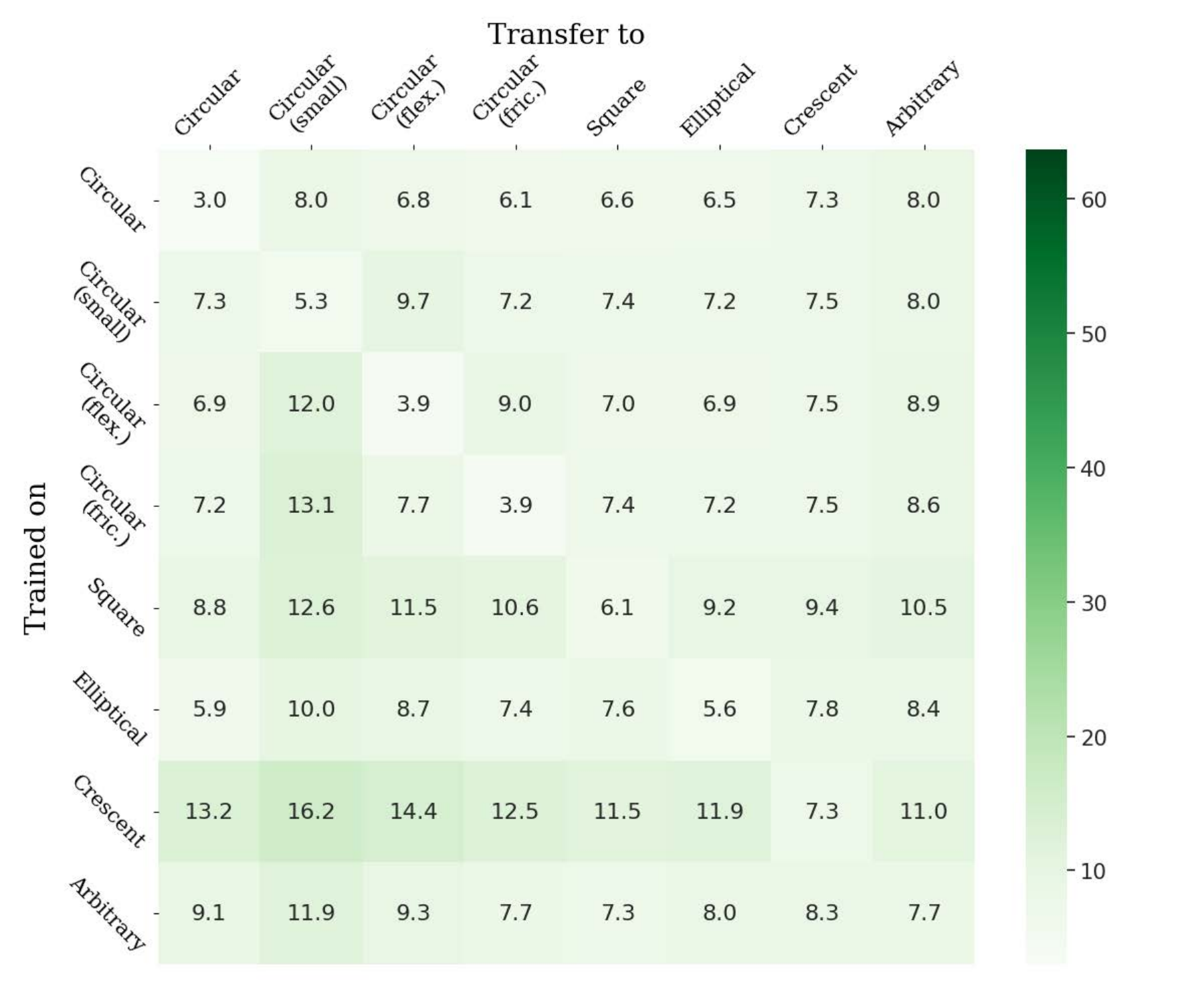} \\ 
\end{tabular}
\caption{Cross transfer pose estimation accuracy between the tested objects when including the initial grasp pose in the state (feature combination 9). Transfer scores include (left) position RMSE in \textit{mm} and (right) orientation errors in \textit{degrees}.}
\label{fig:CMconf9}
\end{figure*}
%%%%%%%%%%%%%%%%%%%%%%%%%%%%%%

%%%%%%%%%%%%%%%%%%%%%%%%%%%%%%
% \begin{figure*}
% \centering
% \begin{tabular}{c}
% \includegraphics[width=1.0\linewidth]{figures/pose_2.png}\\
% \includegraphics[width=1.0\linewidth]{figures/ori_2.png} \\ 
% \end{tabular}
% \caption{Cross transfer pose estimation accuracy with (right column) and without (left column) the initial pose included in the state. Top (green) shows the position RMSE (mm) and bottom (blue) shows the orientation RMSE (deg). X$\and$Y axis represents the object cross-section shape.}
% \vspace{-0.1cm}
% \label{fig:conf}
% \vspace{-0.5cm}
% \end{figure*}
%%%%%%%%%%%%%%%%%%%%%%%%%%%%%%

We next analyze the accuracy with regards to data size. Recall that the total amount of data recorded was $N=92,700$ points. We note that it took approximately 12 hours for automated collection of all data. The accuracy with regards to data size is observed when considering feature combination 7. Hence, we arranged all data sequentially with no shuffling and repeatedly trained the model for a varying portion of the data. The mean prediction accuracy over the test data with the three regression methods can be seen in Figure \ref{fig:portion}. The accuracy improves with the increase of data while the error reaches near saturation at about $60\%$ of the data (corresponds to approximately 7 hours of data collection). The performance provided by the haptics of the hand indicates the ability to perform coarse in-hand manipulation tasks which is experimented in the next sub-section. 

Lastly, we study generalization properties of the LSTM-based observation model for different objects that were not included in the training set. The generalization is analyzed by training a model on one object and evaluating it on test data from the remaining ones. This is repeated for all objects. Figure \ref{fig:CMconf7} shows the cross transfer accuracy across the various objects for feature combination 7. Note that the diagonals of the tables are the object-wise accuracy results presented previously. The study shows that position predictions have a higher transferability than orientation predictions. In general, both position and orientation generalize better when trained over circular objects of varying sizes and textures. Particularly, training over a circular object is sufficient for acquiring a relatively good position accuracy for all objects. The orientation, however, has more difficulty in generalization to more complex objects. On the other hand, Figure \ref{fig:CMconf9} presents the cross transfer accuracy when also including the initial pose to the state (feature combination 9). Clearly, the inclusion of the initial pose significantly improves generalization for both position and orientation predictions, and yields low errors for all objects.

% Figures \ref{fig:CMconf7} and \ref{fig:CMconf9} show the cross transfer performance over different objects with and without initial pose in the state, respectively. 

% This study shows that predicting orientation is more difficult than predicting position. The results show that position predictions have a higher transferability than orientation predictions. Additionally, it generalizes well over circular objects of varying sizes and textures. The inclusion of the initial pose improves generalization for both position and orientation predictions. Comparatively, the model trained on the square object does not transfer well to other objects.

% Results shows higher transfer capabilities of position over orientation predictions. 
% great generalization over circular object with different size and texture
% prediction error is highers with complex object geometry, where orientation is much sensitive the position prediction
% Including the initial pose improve generalization for both position and orientation prediction.
% Interestingly, the model trained on the square object does not transfer well in comparison to other models prediction
% 

% ----------------------------------------------------------------

\subsection{Closed loop planning}

\begin{figure}[]
\centering
  \includegraphics[width=0.9\columnwidth,keepaspectratio]{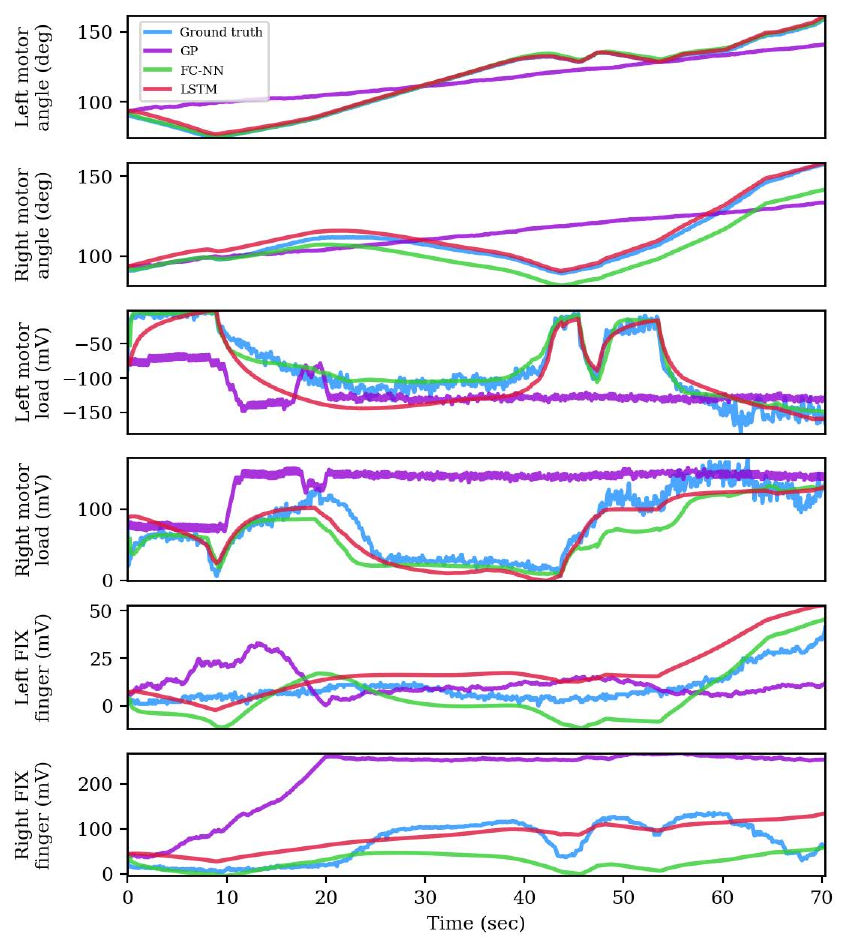} 
  \vspace{-0.6cm}
  \caption{ An example of Conf. 7 feature state transition estimation with three transition models over a test trajectory in $\mathcal{C}$.}
\label{fig:transition_open_loop}
\vspace{-0.3cm}
\end{figure}

\begin{figure}[]
\centering
  \includegraphics[width=\columnwidth,keepaspectratio]{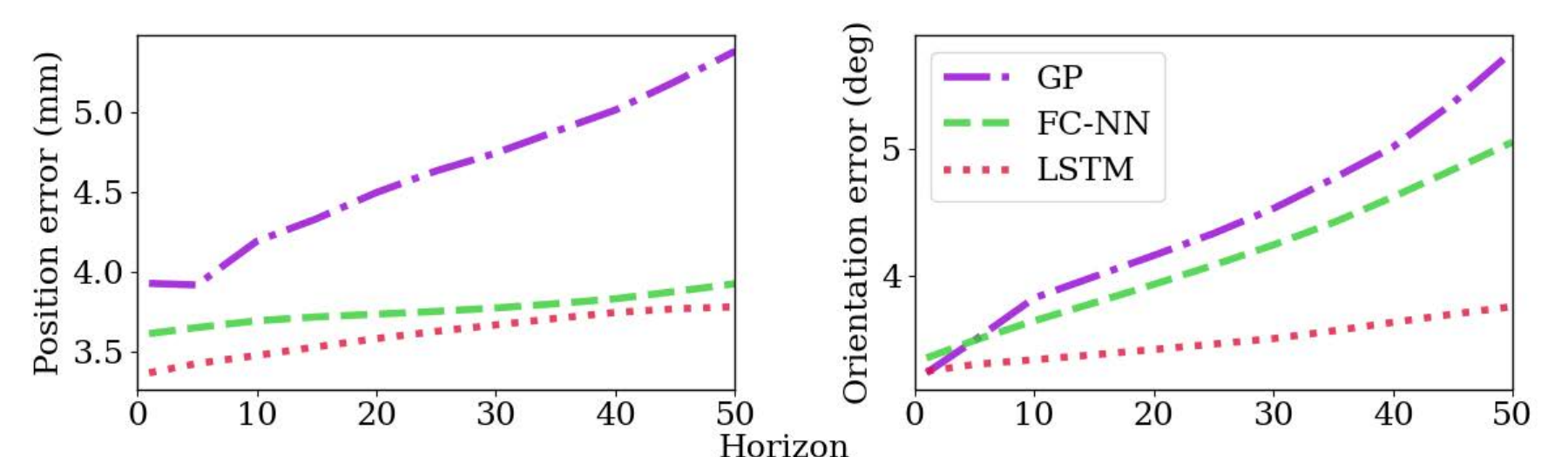} 
  \vspace{-0.6cm}
  \caption{(left) Position and (right) orientation transition accuracy of the three tested models. Predictions are performed in open-loop for an horizon of 50 steps (5.0 seconds).}
\label{fig:transition_horizon}
\vspace{-0.3cm}
\end{figure}

In this section, we test the MPC closed loop control approach discussed in Section \ref{sec:MPC}. We first use the collected data to train transition model \eqref{eq:prediction_state} with feature combination 7. As discussed in Section \ref{sec:transition_model}, the data collected for the observation model is used to train a model $\Delta\tve{x}_t=\tilde{f}_{\phi}(\ve{x}_t,\ve{a}_t)$ and compare between GP, FC-NN and LSTM with $k=2$ past states. The hyper-parameters in $\phi$ were optimized to reduce transition error. We note that the transitions of the model are made in feature space $\mathcal{C}$ while the pose accuracy is computed after using the LSTM observation model presented above. Figure \ref{fig:transition_open_loop} shows a usage example of the three transition models in the feature state space over a test trajectory. Tracking along the ground truth signals is best performed by the LSTM model. Additionally, Figure \ref{fig:transition_horizon} presents the mean error (over the test data) of the three models in an open-loop fashion for an horizon of up to 5.0 seconds. Note that the prediction error is composed of the transition accuracy on top of inherent errors of the observation model. The LSTM model evidently provides better accuracy and is, therefore, used in the following experiments.

%%%%%%%%%%%%%%%%%%%%%%%%%%%%%%
%%%%%%%%%%%%%%%%%%%%%%%%%%%%%%

%%%%%%%%%%%%%%%%%%%%%%%%%%%%%%
%%%%%%%%%%%%%%%%%%%%%%%%%%%%%%

We now observe the performance of the MPC-Critic over a large set of goals in the workspace of the hand. The workspace of the hand is crescent-shaped as illustrated in Figure \ref{fig:traj}. Furthermore, previous work has shown that some goals in the workspace, mostly near the boundaries, are harder to reach due to the risk of getting stuck, slide or drop of the object \cite{Morgan2021c}. The MPC-Critic is implemented according to Algorithm \ref{alg:mpc} with horizon $H=10$ and weight parameters $w_1=0.8$ and $w_2=0.2$. We benchmark the MPC-Critic with three baseline approaches: Open-loop (OL), HS and MPC without critic. For the OL, we incorporate the MPC-Critic in an off-line manner where a set of actions is planned to reach the goal from the mean of all initial grasp locations in the data. Once the object is grasped, the planned actions are rolled-out in open-loop. The HS is implemented as described in Section \ref{sec:HS} where $K_x=K_y=200$. The MPC without critic considers only the first component in the sum of \eqref{eq:cost} with $w_1=1$ and, therefore, aims to shorten the path to the goal. Furthermore, we compare our results to control when visual feedback is available: VS as described in Section \ref{sec:HS} and MPC with a vision-based transition model. For the latter, termed \textit{MPC-Vis}, we use a transition model proposed in \cite{Sintov2019} where the state comprises of the camera perceived pose and actuator torques. MPC-Vis is implemented similar to MPC without critic. The experiment is conducted by randomly sampling 50 poses in the workspace of the hand and attempting to reach them after grasping the object in an arbitrary initial grasp. A roll-out attempt is ended if the object falls or the observation model predicts reaching to the goal. Therefore, a roll-out success is considered if the object reaches within 4 mm to the goal at the end of the attempt. This goal threshold is chosen based to the accuracy limit of combination 7 with LSTM exhibited in the pose estimation results. For the non-visual methods, vision is used solely to estimate and report the true error of reaching a goal. 
%%%%%%%%%%%%%%%%%%%%%%%%%%%%%%
\begin{table}[]
\caption{\small Roll-out results }
\label{tb:control}
\vspace{-0.3cm}
\centering
\begin{tabular}{cc ccc }
\hline
&& Goal error (mm) & Path len. (mm) & Success rate \\ \hline\hline
%  & \multicolumn{1}{c}{\begin{tabular}[c]{@{}c@{}}Goal error\\  (mm)\end{tabular}} & \multicolumn{1}{c}{\begin{tabular}[c]{@{}c@{}} Path length\\ (mm)\end{tabular}} & \begin{tabular}[c]{@{}c@{}}Success rate\\ (\%)\end{tabular} \\ \hline
% \cmidrule[0.4pt](r{0.8em}){3-3}%
% \cmidrule[0.4pt](r{0.8em}){4-4}%
% \cmidrule[0.4pt](r{0.8em}){5-5}%
\multirow{6}{*}{\rotatebox[origin=c]{90}{Circular $15mm$}} & 
OL              & 6.7 $\pm$ 2.0 & 15.4 $\pm$ 9.5  & 40\% \\
& HS              & 6.9 $\pm$ 2.9 & 19.0 $\pm$ 11.8 & 60\% \\
& MPC w/o critic  & 5.0 $\pm$ 2.5 & 36.8 $\pm$ 20.9 & 58\% \\
& MPC-Critic      & 4.7 $\pm$ 2.8 & 41.6 $\pm$ 29.8 & 70\% \\ \cline{2-5}
& VS              & - & 30.6 $\pm$ 16.8 & 73\% \\
& MPC-Vis         & - &  41.2 $\pm$ 15.5 & 82\% \\\hline

\multirow{6}{*}{\rotatebox[origin=c]{90}{Elliptical}} & 
OL              & 7.8 $\pm$ 1.2 & 14.5 $\pm$ 1.0  & 38\% \\
& HS              & 7.1 $\pm$ 3.0 & 17.9 $\pm$ 10.7  & 56\% \\
& MPC w/o critic  & 7.0 $\pm$ 1.8 & 29.4 $\pm$ 14.2  & 60\% \\
& MPC-Critic      & 6.5 $\pm$ 2.3 & 33.5 $\pm$ 20.1  & 74\% \\ \cline{2-5}
& VS              & - & 36.6 $\pm$ 10.5  & 78\% \\
& MPC-Vis         & - & 37.5 $\pm$ 13.4  & 80\% \\ \hline

\multirow{6}{*}{\rotatebox[origin=c]{90}{Crescent}} & 
OL              & 8.2 $\pm$ 1.1 & 16.0 $\pm$ 5.6  & 40\% \\
& HS            & 8.8 $\pm$ 3.1 & 28.7 $\pm$ 12.5  & 60\% \\
& MPC w/o critic & 7.5 $\pm$ 2.0 & 31.2 $\pm$ 12.5  & 58\% \\
& MPC-Critic      & 7.3 $\pm$ 1.0 & 36.5 $\pm$ 22.4  & 70\% \\\cline{2-5}
& VS              & - & 37.5 $\pm$ 14.2  & 70\% \\
& MPC-Vis         & -  & 39.5 $\pm$ 17.2  & 76\% \\ \hline

\end{tabular}
\vspace{-0.6cm}
\end{table}
%%%%%%%%%%%%%%%%%%%%%%%%%%%%%%

\begin{figure*}
\centering
% \begin{tabular}{cccc}
% \includegraphics[width=0.23\linewidth]{figures/a1_new.png} & \includegraphics[width=0.23\linewidth]{figures/a3_new.png} & \includegraphics[width=0.23\linewidth]{figures/a4_new.png} & \includegraphics[width=0.23\linewidth]{figures/a6_new.png} \\
% \includegraphics[width=0.23\linewidth]{figures/a7_new.png} & \includegraphics[width=0.23\linewidth]{figures/a8_new.png} & \includegraphics[width=0.23\linewidth]{figures/a9_new.png} & \includegraphics[width=0.23\linewidth]{figures/a10_new.png} \\
% \end{tabular}
\begin{tabular}{cccc}
\includegraphics[width=0.23\linewidth]{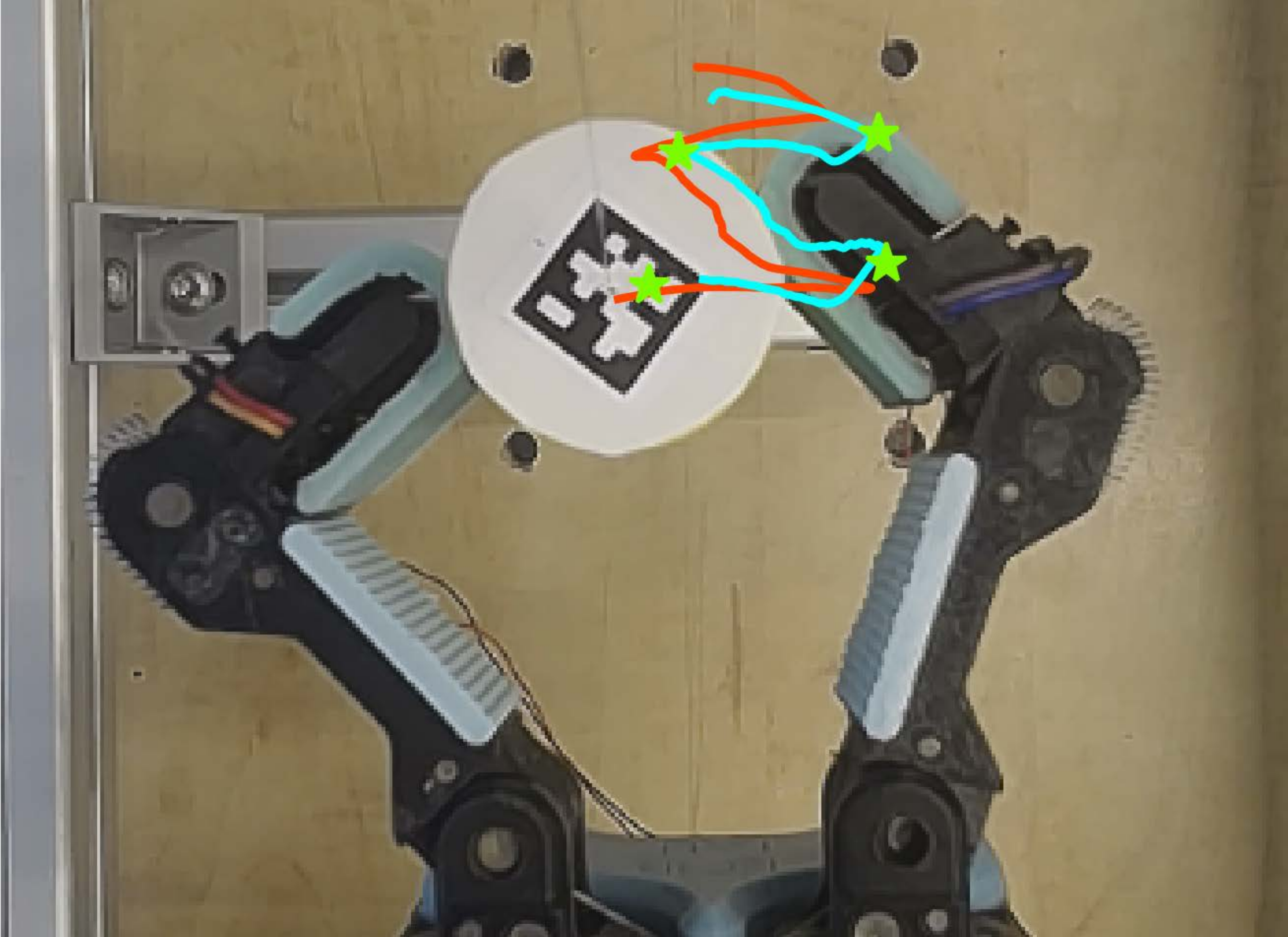} &
\includegraphics[width=0.23\linewidth]{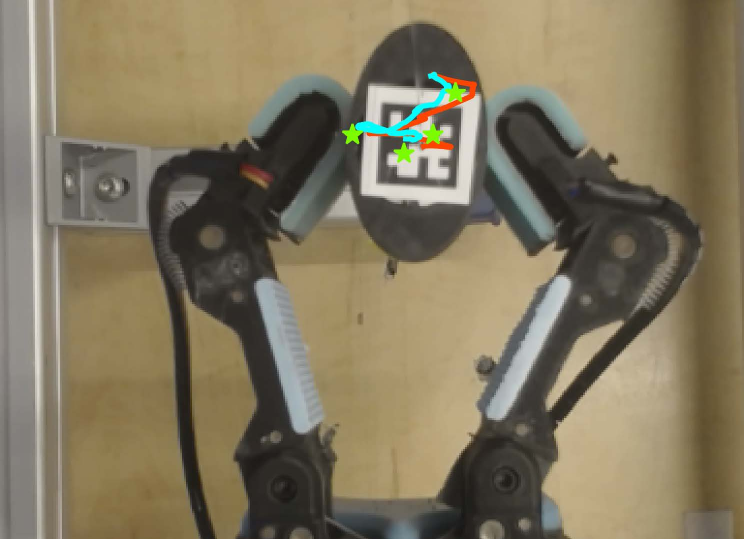} & \includegraphics[width=0.23\linewidth]{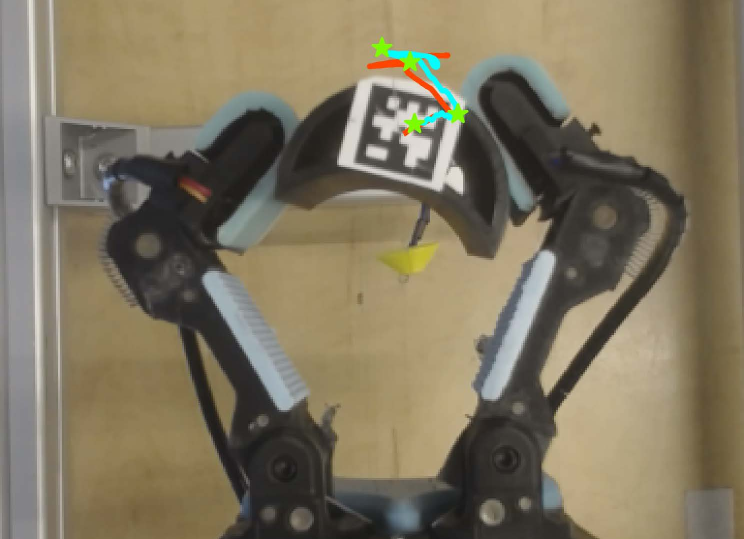} & 
\includegraphics[width=0.23\linewidth]{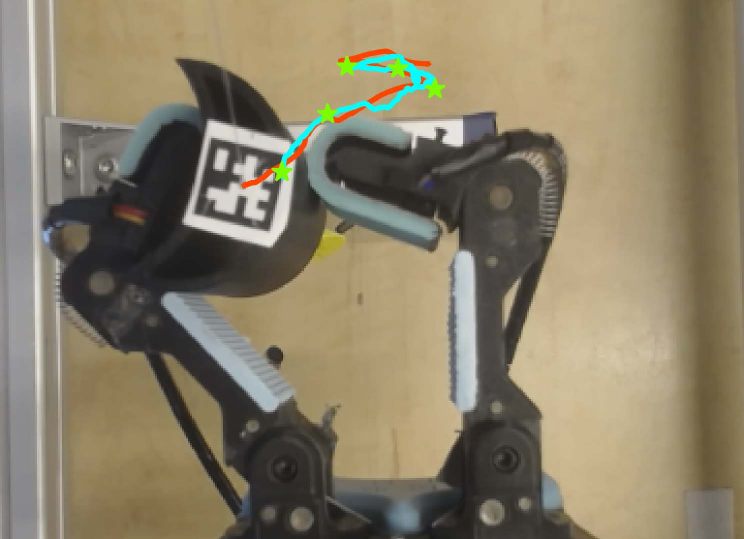} \\
\end{tabular}
% % \includegraphics[width=0.45\linewidth]{figures/a7.PNG} &
\caption{Four roll-out examples where the circular, elliptical and crescent objects are manipulated between several intermediate goals (green markers) using the MPC-Critic. The rolled-out path and position estimation are seen in red and cyan curves, respectively.}
\vspace{-0.1cm}
\label{fig:test}
\vspace{-0.3cm}
\end{figure*}

Table \ref{tb:control} presents averaged results for roll-outs with all methods of the circular (15mm), elliptical and crescent objects. The table shows the mean distance to the goal over all trials after finishing the roll-outs. We first compare between the four non-visual methods. MPC-Critic exhibits the lowest error. In addition, the results show that MPC-Critic is able to reach farther goals with higher success rate. The critic has sparse data in regions which are harder to cross due to high probability of dropping the object or actuator overload. Hence, the critic has lower accuracy in those regions such that the MPC-Critic inherently avoids them. In addition, the goals were randomly sampled and the system is not guaranteed to be able to reach them. Figure \ref{fig:test} shows roll-out demonstrations with the MPC-Critic where the object must move between several intermediate goals. The accuracies of the visual-based methods, MPC-Vis and VS, are equal to the success tolerance and, therefore, not reported. However and as expected, they exhibit slightly higher success rates than MPC-Critic due to more accurate pose feedback. Nevertheless, the results validate the proposed approach and show feasible manipulations with solely haptic feedback. 

We have also conducted a peg-in-a-hole demonstration in which the objects are manipulated within a confined cabinet into slots. The positions of the slots relative to the hand are assumed to be known. While not in our scope, further work is required in order to sense the target hole, given the general location, using the available haptic feedback. Images of the demonstration are seen in Figure \ref{fig:hand}. Videos of the demonstration and other experiments are included in the supplementary video.

%%%%%%%%%%%%%%%%%%%%%%%%%%%%%%
% \begin{figure*}
% \centering
% \begin{tabular}{cc}
% \includegraphics[height=1.9cm]{figures/hand_new2.png}  &
% \includegraphics[height=1.9cm]{figures/path.png} \\
% (a) & (b) 
% \end{tabular}
% \caption{(a) Examples of roll-outs to two goals using the MPC-Critic and, (b) the true and predicted paths of the object. }
% \label{fig:test}
% \vspace{-0.5cm}
% \end{figure*}
%%%%%%%%%%%%%%%%%%%%%%%%%%%%%%

%%%%%%%%%%%%%%%%%%%%%%%%%%%%%%

\section{Conclusions}
In this letter, we have investigated the ability to track and manipulate an object within a two-finger underactuated hand solely based on haptics without relying on visual perception. We explored the state representation of an observation model using the kinesthetic information available to the hand along with novel low-cost tactile fingers. This observation is valuable since vision is often limited, i.e., working in an occluded environment. Adding an initial pose glance at the moment of grasping has shown to provide an additional accuracy improvement. Additionally, we proposed a data-driven MPC approach which utilized learnt observation and transition models to reach desired goals within the workspace of the hand. The MPC approach reasons about the accuracy of the observation model using a critic and attempts to avoid erroneous regions. Results have shown the ability to estimate the pose of several test objects and manipulate them to desired goals with less than 5 mm accuracy. Moreover, generalization to various object was shown feasible. The results validate that in-hand manipulation is feasible with solely haptic feedback. However, the accuracy achieved by the system may not be sufficient in more complex tasks such inserting a key into a lock. These may require higher resolution hardware. An interesting extension to the work may include more intelligent exploration methods that can increase data quality and be more sample-efficient. %Moreover, we would consider to 
%explore general approaches to deal with various objects and 
%extend our analysis to spatial in-hand manipulation tasks.

\bibliographystyle{IEEEtran}
\bibliography{ref}

% \newpage

\section{Biography}
% If you have an EPS/PDF photo (graphicx package needed), extra braces are
%  needed around the contents of the optional argument to biography to prevent
%  the LaTeX parser from getting confused when it sees the complicated
%  $\backslash${\tt{includegraphics}} command within an optional argument. (You can create
%  your own custom macro containing the $\backslash${\tt{includegraphics}} command to make things
%  simpler here.)
 
\vspace{-1.2cm}

% \bf{If you include a photo:}\vspace{-33pt}
\begin{IEEEbiography}[{\includegraphics[width=1in,height=1.25in,clip,keepaspectratio]{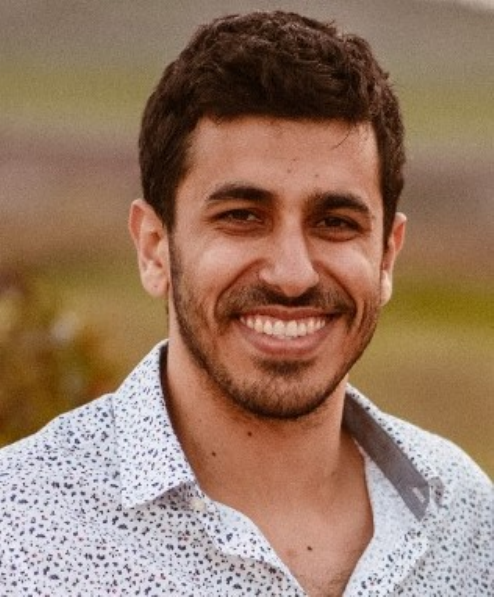}}]{Osher Azulay}
received his B.Sc. and M.Sc. in Mechanical Engineering from Ben-Gurion University at 2019 and 2020, respectively. He is currently a Ph.D. student at the School of Mechanical Engineering at Tel-Aviv University. His research interests include in-hand manipulations, robot learning and tactile sensing.
\end{IEEEbiography}

\vspace{-1.2cm}

\begin{IEEEbiography}[{\includegraphics[width=1in,height=1.25in,clip,keepaspectratio]{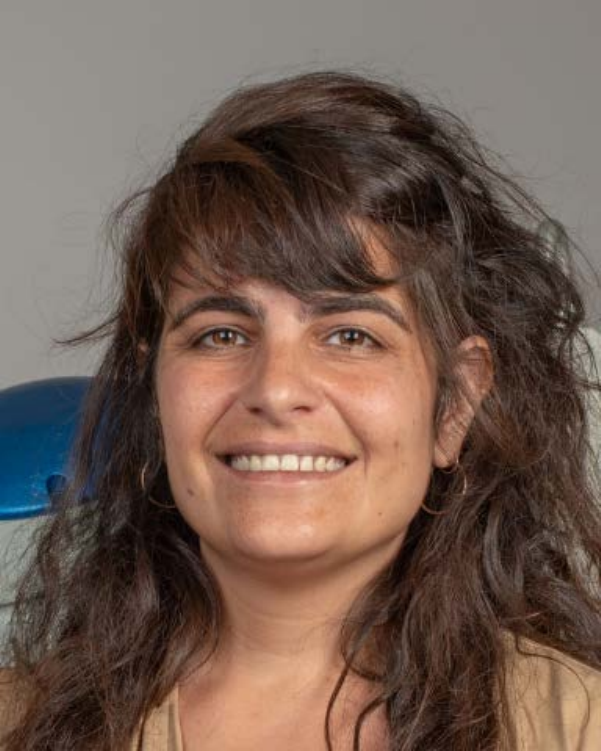}}]{Inbar Ben-David}
received her B.Sc. in Mechanical Engineering from Ben-Gurion and MBA in Business Administration from the College of Management Academic Studies at 2017 and 2019, respectively. She is currently a M.Sc student at the School of Mechanical Engineering at Tel-Aviv University. Her research interests include robot design and genetic optimization.
\end{IEEEbiography}

\vspace{-1.2cm}

\begin{IEEEbiography}[{\includegraphics[width=1in,height=1.25in,clip,keepaspectratio]{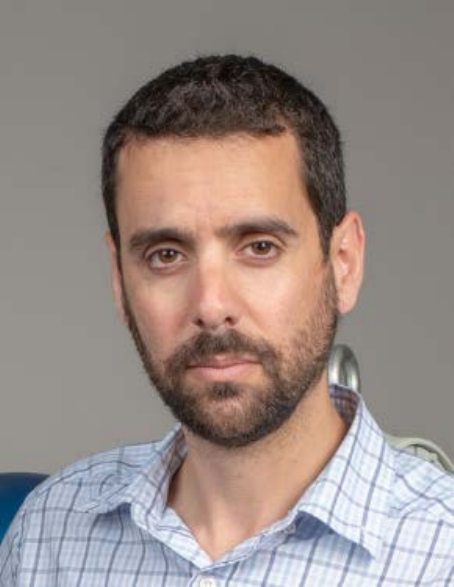}}]{Avishai Sintov}
received his B.Sc. M.Sc. and Ph.D. in 2008, 2012 and 2016, respectively, from the department of Mechanical Engineering at Ben-Gurion University of the Negev. He is currently an assistant professor at the School of Mechanical Engineering at Tel-Aviv University. Previously, he was a post-doc fellow at the University of Illinois at Urbana-Champaign (2016-2018) and Rutgers University (2018-2019). His research interests include Human-Robot Collaboration dynamic manipulations, grasping, regrasping, constrained systems, machine learning and robot design.
\end{IEEEbiography}

% \bf{If you will not include a photo:}\vspace{-33pt}
% \begin{IEEEbiographynophoto}{John Doe}
% Use $\backslash${\tt{begin\{IEEEbiographynophoto\}}} and the author name as the argument followed by the biography text.
% \end{IEEEbiographynophoto}

\vfill

\end{document}